\documentclass[lettersize,journal]{IEEEtran}

\usepackage{amsmath,amsfonts}
\def\BibTeX{{\rm B\kern-.05em{\sc i\kern-.025em b}\kern-.08em
    T\kern-.1667em\lower.7ex\hbox{E}\kern-.125emX}}
\usepackage{balance}

\usepackage[numbers]{natbib}
\usepackage{multicol}
\usepackage[bookmarks=true]{hyperref}
\usepackage{graphics} % for pdf, bitmapped graphics files
\usepackage{epsfig} % for postscript graphics files
\usepackage{times} % assumes new font selection scheme installed
\usepackage{amssymb}  % assumes amsmath package installed
\usepackage{hyperref} % for url links
\usepackage{xcolor} % for text in colors
\usepackage{arydshln} % dashlines in tables
\usepackage{color}
\usepackage{float}
\usepackage{multirow}
\usepackage{booktabs}
\usepackage{tikz}
\usepackage{wrapfig}
\usepackage{units}
\usepackage{bm}
\usepackage{tabularx}
\usetikzlibrary{calc ,math, positioning, intersections, decorations, external, plotmarks, shapes}
\usepackage{pgfplots}
\usepackage{pgfplotstable}
\usepgfplotslibrary{statistics}
\pgfplotsset{compat=1.16}
\usepackage[skip=0pt,font=footnotesize]{caption}
\usepackage{subcaption}

\newcommand{\rebuttal}[1]{{\color{black} #1}}

\newcommand{\camera}[1]{{\color{black} #1}}

\newcommand{\wfr}[0]{\ensuremath{\mathcal{W}}} % world frame
\newcommand{\bfr}[0]{\ensuremath{\mathcal{B}}} % body frame
\newcommand{\cfr}[0]{\ensuremath{\mathcal{C}}} % C-frame
\newcommand{\ifr}[0]{\ensuremath{\mathcal{I}}} % I-frame

\newcommand\norm[1]{\left\lVert#1\right\rVert}

\newcolumntype{C}{>{\centering\arraybackslash}X}
\newcolumntype{P}[1]{>{\centering\arraybackslash}p{#1}}

\definecolor{matlab1}{rgb}{0.00000,0.44700,0.74100}%
\definecolor{matlab2}{rgb}{0.85000,0.32500,0.09800}%
\definecolor{matlab3}{rgb}{0.92900,0.69400,0.12500}%
\definecolor{matlab4}{rgb}{0.49400,0.18400,0.55600}%
\definecolor{matlab5}{rgb}{0.4660, 0.6740, 0.1880}%
\definecolor{matlab6}{rgb}{0.3010, 0.7450, 0.9330}%
\definecolor{matlab7}{rgb}{1.0, 0.0, 0.0}%

\def\method{HDVIO2.0}
\def\ourMethod{HDVIO2.0 (Ours)}

\def\methodPrev{HDVIO}

\definecolor{somegray}{rgb}{0.5, 0.5, 0.5}
\newcommand{\darkgrayed}[1]{\textcolor{somegray}{#1}}
\makeatletter
\newcommand*\titleheader[1]{\gdef\@titleheader{#1}}
\AtBeginDocument{%
  \let\st@red@title\@title
  \def\@title{%
    \vskip-2em
    \bgroup\normalfont\large\centering\@titleheader\par\egroup
    \vskip0.5em\st@red@title}
}
\makeatother

\titleheader{\darkgrayed{This paper has been accepted for publication at the
Transactions on Robotics (T-RO) 2025}}

\title{HDVIO2.0: Wind and Disturbance Estimation with Hybrid Dynamics VIO}
\begin{document}

\author{Giovanni Cioffi, Leonard Bauersfeld, Davide Scaramuzza
\thanks{The authors are with the Robotics and Perception Group, Department of Informatics, University of Zurich, Switzerland, \protect\url{https://rpg.ifi.uzh.ch}.\newline
This work was supported by the European Union’s Horizon Europe Research and Innovation Programme under grant agreement No. 101120732
(AUTOASSESS) and the European Research Council (ERC) under grant agreement No. 864042 (AGILEFLIGHT).}}

% \markboth{Journal of \LaTeX\ Class Files,~Vol.~18, No.~9, September~2020}%
% {How to Use the IEEEtran \LaTeX \ Templates}

\maketitle

\begin{abstract}
Visual-inertial odometry (VIO) is widely used for state estimation in autonomous micro aerial vehicles using onboard sensors. 
Current methods improve VIO by incorporating a model of the translational vehicle dynamics, yet their performance degrades when faced with low-accuracy vehicle models or continuous external disturbances, like wind. 
Additionally, incorporating rotational dynamics in these models is computationally intractable when they are deployed in online applications, e.g., in a closed-loop control system.
We present \method, which models full 6-DoF, translational and rotational, vehicle dynamics and tightly incorporates them into a VIO system with minimal impact on the runtime.
\method\ builds upon the previous work, \methodPrev, and addresses these challenges through a hybrid dynamics model combining a point-mass vehicle model with a learning-based component, with access to control commands and IMU history, to capture complex aerodynamic effects.
The key idea behind modeling the rotational dynamics is to represent them with continuous-time functions.
\method\ leverages the divergence between the actual motion and the predicted motion from the hybrid dynamics model to estimate external forces as well as the robot state.
Our system surpasses the performance of state-of-the-art methods in experiments using public and new drone dynamics datasets, as well as real-world flights in winds up to 25 km/h.
%
% Moreover, we include \method\ in a closed-loop control system and show autonomous flights of a racer drone. 
%
Unlike existing approaches, we also show that accurate vehicle dynamics predictions are achievable without precise knowledge of the vehicle state.

\end{abstract}


% SECTIONS
\section*{Supplementary Material}\label{sec:SupplementaryMaterial}

\textbf{Video}: \url{https://youtu.be/wUaEp0YGpDM}

\textbf{Code}:\url{https://github.com/uzh-rpg/hdvio2.0}
\section{Introduction}\label{sec:Introduction}
\IEEEPARstart{V}{isual-inertial} odometry (VIO) is the standard method for state estimation in consumer and inspection drones. 
To enhance the performance of VIO systems, several recent approaches have proposed tightly integrating drone dynamics into the VIO pipeline~\cite{nisar2019vimo, ding2021vid, chen2022visual, fourmy2021contact, kang2023view}.
Incorporating system dynamics into the VIO framework provides additional information, enabling the system to distinguish between motion resulting from actuation and motion caused by external perturbations. 
This integration improves pose estimation accuracy and allows for the estimation of external forces acting on the drone.

While effective in many scenarios, state-of-the-art methods face significant performance degradation in cases of large model mismatches (e.g., high speeds, systematic noise in actuation inputs) or persistent external disturbances like wind. 
These issues arise because existing methods rely on simplifying assumptions—such as neglecting aerodynamic drag and assuming zero-mean noise in system dynamics—that fail to hold under such conditions. 
Directly incorporating high-fidelity dynamics models~\cite{bauersfeld2021neurobem,sun2019quadrotor} into a VIO pipeline can be counterproductive because these models require the drone state as input (typically velocity and attitude).
This situation can create a compounding effect, where errors in the VIO output propagate through the dynamics model, and, in turn, further impact the VIO.

Overcoming these challenges is essential for deploying model-based VIO estimators in applications where aerodynamic effects play a significant role, such as fast flights~\cite{bauersfeld2022rangeestimates}, operations in windy conditions~\cite{connell2022neuralfly}, or scenarios with modeling inaccuracies~\cite{kaufmann2023champion}.
\begin{figure}
    \centering
    \includegraphics[width=\linewidth]{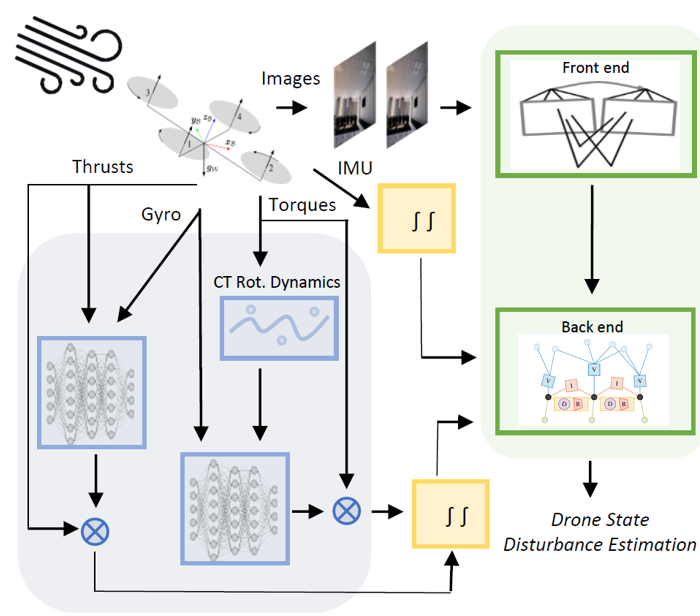}
    \vspace*{6pt}
    \caption{\method\ estimates the robot's state and external disturbances using visual, inertial, and dynamics measurements. Notably, it models the robot's dynamics by combining a simple physical model with learning-based components.}
    \vspace*{-12pt}
    \label{fig:fig1}
\end{figure}
The state-of-the-art methods VIMO~\cite{nisar2019vimo} and VID~\cite{ding2021vid} incorporate the translational drone dynamics into an optimization-based VIO framework~\cite{qin2018vins} through a residual term derived from the propagation of a point-mass dynamics model.
The residual term is formulated based on the preintegration theory~\cite{forster2016manifold}, which requires separating the measurements (namely, the control inputs) from the states.
While this formulation is straightforward for translational dynamics, extending it to rotational dynamics is not trivial.
The simplified dynamics model neglects aerodynamic effects, treating drag as part of the external force estimate. 
Additionally, potential systematic offsets in actuation inputs (e.g., miscalibrations such as incorrect rotor lift coefficients) are interpreted as accelerometer biases, introducing errors into the inertial residuals and reducing motion estimation accuracy.

Improving the drone dynamics model within the VIO estimator is key to addressing these limitations.

\subsection*{Contribution}
We present \method, the first VIO pipeline incorporating the 6-DoF drone dynamics using a hybrid model that combines a physical model with a learning-based component, see Fig~\ref{fig:fig1}. 
Unlike prior methods~\cite{bauersfeld2021neurobem,sun2019quadrotor}, our learned dynamics model predicts residual aerodynamic forces without requiring the drone state (i.e., attitude and velocity) as input. 
Instead, it uses a temporal convolutional network (TCN)~\cite{oord2016wavenet} that takes control commands and gyroscope measurements as input. 
Our hybrid model is integrated into an optimization-based VIO framework~\cite{Forster17troSVO, leutenegger2015keyframe}, leveraging preintegration theory~\cite{forster2016manifold} to efficiently compute and optimize the dynamics residuals alongside monocular camera and IMU residuals.
\method\ extends the previous work, \methodPrev~\cite{cioffi2023hdvio}, by integrating the rotational dynamics in the drone model.
We represent the drone's angular velocity as a continuous-time function using B-splines, which effectively capture the drone’s dynamics~\cite{cioffi2022continuous}.
The B-spline is optimized by minimizing the difference between its first derivative and the angular acceleration derived from the drone's rotational model, which combines torque commands and learned residual torques.
Angular velocities sampled from the B-spline representation of the drone's rotational dynamics model are preintegrated into a residual term that can be directly included in the VIO.
\method\ employs two TCNs: one predicts residual thrust using thrust commands and gyroscope measurements (as in HDVIO), while the other predicts residual torque based on torque commands and gyroscope measurements—a novel contribution.

We evaluate \method\ against the same VIO system without the proposed hybrid dynamics model, and the state-of-the-art systems, VIMO~\cite{nisar2019vimo} and VID~\cite{ding2021vid}. 
On public datasets, Blackbird~\cite{antonini2020blackbird} and VID~\cite{zhang2022visual}, we show that \method\ surpasses state-of-the-art methods.
% improving robot state and external force estimation by up to x\% and x\%, respectively. 
%
In wind field experiments, we show that \method\ accurately predicts the wind force, again outperforming the baselines.
On the NeuroBEM dataset~\cite{bauersfeld2021neurobem}, our learned dynamics model demonstrates competitive performance with existing aerodynamics models.
Notably, it is the first data-driven dynamics model to predict forces without requiring the vehicle state, i.e., linear velocity and attitude, as input.
Additionally, to the best of our knowledge, our learning-based model is the first data-driven dynamics model that trains without requiring ground-truth force measurements, relying solely on position, velocity, and orientation supervision signals.
This eliminates the need for motion-capture systems, as pose estimates obtained from offline structure from motion-based systems~\cite{schoenberger2016sfm, cioffi2022continuous} are sufficiently accurate for training.

By providing accurate state and external force estimates, we believe that \method\ advances the deployment of autonomous drones in safety-critical applications, such as disaster site surveying and air transport, which currently depend on human pilots.
\section{Related Work}\label{sec:RelatedWork}
The related work on visual-inertial odometry (VIO) with external force estimation can be categorized into \textit{loosely-coupled} and \textit{tightly-coupled} methods. 
Loosely-coupled approaches~\cite{tomic2014unified, yuksel2014nonlinear, ruggiero2014impedance, mckinnon2016unscented, augugliaro2013admittance, tagliabue2017collaborative,abeywardena2014model, walti2024lumped} estimate external forces independently from motion estimation, while tightly-coupled methods~\cite{nisar2019vimo, ding2021vid, fourmy2021contact} jointly estimate both the robot's motion and external perturbations.  

\subsection{Loosely-Coupled Methods}
Early loosely-coupled methods~\cite{tomic2014unified, yuksel2014nonlinear, ruggiero2014impedance} rely on deterministic force and torque observers derived from the robot's dynamics model. 
These approaches assume access to accurate state estimates from separate estimators.  
Later, probabilistic methods~\cite{mckinnon2016unscented, augugliaro2013admittance, tagliabue2017collaborative, abeywardena2014model} improved accuracy by incorporating sensor noise.
These methods utilize the Extended Kalman Filter (EKF)~\cite{augugliaro2013admittance, abeywardena2014model, 10342358} or the Unscented Kalman Filter (UKF)~\cite{mckinnon2016unscented,tagliabue2017collaborative}. 
For example, the work in~\cite{chen2022visual} uses a quadrotor model to enhance an EKF-based VIO estimator~\cite{mourikis2007multi} for simultaneous system identification and state estimation. 
This work highlights that decoupling state estimation from dynamics measurements is optimal in the presence of high noise. 
In~\cite{tagliabue2020touch}, a UKF estimates external disturbances like wind and human interactions using outputs from a neural network that processes airflow sensor data and motion capture measurements. 

While effective under high signal-to-noise ratio conditions, loosely-coupled methods neglect the correlation between estimated variables and their noise characteristics, resulting in reduced performance when using noisy sensors.  

\subsection{Tightly-Coupled Methods}
Tightly-coupled methods address this limitation by jointly estimating robot motion and external perturbations. 
VIMO~\cite{nisar2019vimo} integrates robot dynamics into an optimization-based VIO framework~\cite{qin2018vins}. 
It introduces a residual term representing translational motion constraints derived from robot dynamics, including external forces, using IMU preintegration theory~\cite{forster2016manifold}.
This method preintegrates high-rate thrust inputs into residuals between consecutive camera frames.
External forces are modeled as zero-mean Gaussian variables to account for their unknown dynamics.
VIMO is the first work that enables the simultaneous estimation of external forces and robot states. 
Multiple extensions to VIMO exist.
The work in~\cite{moeini2024visual} extends VIMO with a disturbance observer for constant force estimation.
The disturbance observer allows the system to differentiate between the constant external force and the accelerometer bias.
VID-Fusion~\cite{ding2021vid} extends VIMO by modifying the external force model, where the mean of the Gaussian distribution is based on the average difference between accelerometer and thrust measurements within the preintegration window.
This model of the external force allows VID-Fusion to estimate constant loads attached to the drone.
We employ this external force model in~\method.
The VIMO framework is general and can be employed for any type of robot. 
In fact, extensions of VIMO for legged robots are proposed in~\cite{fourmy2021contact, kang2023view}.

However, as discussed in Sec.~\ref{sec:Introduction}, VIMO and all its extensions struggle with continuous external forces or model mismatches.  

\subsection{Drone Dynamics Modeling}
Accurate dynamics modeling is critical for \method. 
Prior methods assume access to the vehicle state, which is unsuitable for VIO pipelines as it introduces a compounding effect that propagates errors in the dynamics model to the VIO and vice versa.
For completeness, a brief review of quadrotor modeling literature is presented.  
Basic models treat quadrotors as rigid bodies with linear mass and inertia dynamics, exerting force in the body-z direction while neglecting or simplifying (assuming it linear) aerodynamic drag~\cite{furrer2016rotors, song2020flightmare, shah2018airsim, meyer2012comprehensive}. 
First principles can be used to refine these basic models, resulting in blade-element momentum (BEM) theory~\cite{bauersfeld2021neurobem, gill2017propeller, hoffmann2007quadrotor, orsag2012influence}.
Pure data-driven models~\cite{sun2019quadrotor} have gained traction due to the complexity of quadrotor aerodynamics and have shown superior performance compared to first-principles-based methods. 
The state-of-the-art model, NeuroBEM~\cite{bauersfeld2021neurobem}, combines a physical model with a learning-based component, outperforming previous methods. 
This hybrid modeling approach inspired our use of a learned component in \method\ to enhance drone dynamics modeling.

\rebuttal{\method\ extends \methodPrev\ by modeling the quadrotor's rotational dynamics using a continuous-time representation of angular velocities, implemented via B-splines for computational efficiency. 
Alternatively, Gaussian Processes (GPs) have been used as CT representations in prior work~\cite{le2021continuous, le2023continuous, li2024asynchronous} to derive preintegration terms based on IMU measurements. 
In~\cite{le2021continuous, le2023continuous}, the 6-DoF sensor pose is modeled using six independent GPs. 
These GPs are optimized to fit IMU measurements, and preintegration terms for orientation, velocity, and position are obtained either by (i) sampling angular velocities and linear accelerations followed by numerical integration~\cite{le2021continuous}, or (ii) applying linear integration operators directly to the GPs~\cite{le2023continuous}.
The method proposed in~\cite{le2023continuous} is applied in an event-based odometry system in~\cite{li2024asynchronous}.
While GPs provide accurate results, their high computational cost makes them impractical for real-time use in \method.}
\section{Methodology}\label{sec:Methodology}
This section outlines our visual-inertial-hybrid drone dynamics odometry algorithm.
We begin by defining the notation used throughout the paper and describing the drone dynamics.
%
% While our derivation focuses on a quadrotor platform, the proposed approach is adaptable to other robotic systems.
%
Next, we formulate the estimation problem.
Following, we provide a concise derivation of the dynamics residual term, which is based on the preintegration theory~\cite{forster2016manifold}.
Finally, we introduce our learning-based module of drone dynamics.

\subsection{Notation}
In this paper, scalars are represented using non-bold notation~$[s, S]$, vectors are denoted in lowercase bold~$\bm{v}$, and matrices are expressed in uppercase bold~$\bm{M}$. 
World $\wfr$, Body $\bfr$, IMU $\ifr$, and camera $\cfr$ frames are defined with an orthonormal basis, such as $\{\bm{x}^\wfr, \bm{y}^\wfr, \bm{z}^\wfr\}$.
The $\bfr$ frame is positioned at the quadrotor's center of mass, and for simplicity, the IMU frame $\ifr$ is assumed to coincide with $\bfr$.
The notation $(\cdot)^{\wfr}$ is used to indicate quantities expressed in the world frame, and similar notation is applied to other reference frames.  
At time $t_k$, the position, orientation, and linear velocity of $\bfr$ relative to $\wfr$ are denoted as $\bm{p}_{\bfr_{k}}^{\wfr} \in \mathbb{R}^3$, $\bm{R}_{\bfr_{k}}^{\wfr} \in \mathbb{R}^{3 \times 3}$ (a member of the rotation group $SO(3)$), and $\bm{v}_{\bfr_{k}}^{\wfr} \in \mathbb{R}^3$, respectively.
%
% The quadrotor body rate is denoted by $\bm{\omega}^{\bfr_{k}}$.
%
The unit quaternion representation of $\bm{R}_{\bfr_{k}}^{\wfr}$ is given as $\bm{q}_{\bfr_{k}}^{\wfr}$.   
The cross product of two vectors is denoted by $\times$.
The quaternion product is denoted by $\otimes$.
The gravity vector in the world frame is denoted by $\bm{g}^\wfr$.
The accelerometer model is: $\hat{\bm{a}}^{\bfr_{k}} = \bm{a}^{\bfr_{k}} + \bm{b}_{a_{k}} + \bm{n}_{a}$, where the noise is modeled as additive Gaussian noise $\bm{n}_{a} \thicksim \mathcal{N}(0, \bm{\sigma}_{a}^2)$ and the bias as a random walk $\bm{\dot{b}}_{a_{k}} = \bm{n}_{b_{a}}$, with $\bm{n}_{b_{a}} \thicksim \mathcal{N}(0, \bm{\sigma}_{b_{a}}^2)$.
The gyroscope model is: $\hat{\bm{\omega}}^{\bfr_{k}}_g = \bm{\omega}^{\bfr_{k}}_g + \bm{b}_{\omega_{gk}} + \bm{n}_{\omega_g}$, where the noise is modeled as additive Gaussian noise $\bm{n}_{\omega_g} \thicksim \mathcal{N}(0, \bm{\sigma}_{\omega_g}^2)$ and the bias as a random walk $\bm{\dot{b}}_{\omega_{gk}} = \bm{n}_{b_{\omega_g}}$, with $\bm{n}_{b_{\omega_g}} \thicksim \mathcal{N}(0, \bm{\sigma}_{b_{\omega_g}}^2)$.
We indicated noisy measurements using the symbol $\hat{\cdot}$.

\subsection{Quadrotor Dynamics}
\rebuttal{The quadrotor is modeled as a 6 degree-of-freedom rigid body with mass $m$ and a diagonal moment of inertia matrix $\bm{J} = diag(J_x, J_y, J_z)$.}
The dynamics governing the position, velocity, and orientation of the quadrotor platform are described by the following equations:
\begin{align}
    \dot{\bm{p}}^{\wfr}_{\bfr_k} & = \bm{v}^{\wfr}_{\bfr_k} \nonumber \\
    \dot{\bm{v}}^{\wfr}_{\bfr_k} & = \bm{R}^{\wfr}_{\bfr_k} (\bm{f}^{\bfr}_{t_k} + \bm{f}^{\bfr}_{res_k} + \bm{f}^{\bfr}_{e_k}) + \bm{g}^{\wfr} \nonumber \\
    \dot{\bm{q}}^{\wfr}_{\bfr_k} & = \frac{1}{2} \bm{q}^{\wfr}_{\bfr_k} \otimes [0, \bm{\omega}^{\bfr^{\top}}_k]^{\top} \nonumber \\
    \dot{\bm{\omega}}^{\bfr}_k & = \bm{J}^{-1} (\bm{\tau}^{\bfr}_{k} + \bm{\tau}^{\bfr}_{res_k} - \bm{\omega}^{\bfr}_k \times \bm{J} \bm{\omega}^{\bfr}_k),
    \label{eq:drone_model}
\end{align}
where $\bm{f}^{\bfr}_{t_k} = [0, 0, T_k]^{\top}$ represents the mass-normalized collective thrust, $\bm{f}^{\bfr}_{e_k}$ denotes the external force acting on the quadrotor, $\omega^{\bfr}_k$ is the quadrotor's body rate, and $\bm{\tau}^{\bfr}_{k}$ is the torque produced by the propellers.
\rebuttal{We define the mass-normalized collective thrust measurement model as: $\hat{\bm{f}}^{\bfr}_{t_k} = \bm{f}^{\bfr}_{t_k} + \bm{n}_{f_{t}}$, where $\bm{n}_{f_{t}} \thicksim \mathcal{N}(0, \bm{\sigma}_{f_{t}}^2)$ is a zero-mean gaussian noise to account for uncertainty in the force direction.}
For conciseness, we will refer to the mass-normalized collective thrust as simply thrust hereafter.
\rebuttal{Similarly, the torque measurement model is: $\hat{\bm{\tau}}^{\bfr}_{k} = \bm{\tau}^{\bfr}_{k} + \bm{n}_{\tau}$, where $\bm{n}_{\tau} \thicksim \mathcal{N}(0, \bm{\sigma}_{\tau}^2)$ is a zero-mean gaussian noise.}
%
% The torque measurement $\hat{\bm{\tau}}^{\bfr}_{k}$ is used to optimize the B-spline representing the quadrotor body rates as described in Sec.~\ref{sec:ct_quad_dynamics}.
%
The body rate measurement model is: $\hat{\bm{\omega}}^{\bfr}_k = \bm{\omega}^{\bfr}_k + \bm{n}_{\omega}$, where $\hat{\bm{\omega}}^{\bfr}_k$ is sampled from the B-spline, see Sec.~\ref{sec:ct_quad_dynamics}, and $\bm{n}_{\omega}$ is a noise value that accounts for uncertainty in the B-spline fitting process.
To account for aerodynamic effects and unknown systematic noise in the inputs, residual terms $\bm{f}^{\bfr}_{res_k}$ and $\bm{\tau}^{\bfr}_{res_k}$ are introduced.
The external force $\bm{f}^{\bfr}_{e_k}$ is modeled as a random variable following a Gaussian distribution obtained by computing the difference between acceleration and thrust measurements as proposed in~\cite{ding2021vid}.
Modeling the external force in this manner enables the estimator to differentiate between the slowly varying accelerometer bias and external forces, which may arise from incidental disturbances or constant loads, such as an external mass attached to the drone.

The dynamics motion constraints, c.f. Section~\ref{sec:dynamics_residuals}, are derived using the preintegration theory~\cite{forster2016manifold}.
\rebuttal{The preintegration theory allows for efficient integration of these constraints in the VIO optimization-based backend.}
\rebuttal{To apply the preintegration theory, a measurement model of the form $\bm{y} = f(\bm{x})$ is required in which the optimization variables $\bm{x}$ can be separated from the measurements $\bm{y}$, and $f(\cdot)$ describes the relationship between $\bm{x}$ and $\bm{y}$.}
\rebuttal{The quadrotor rotational dynamics (last two equations of Eq.~\ref{eq:drone_model}) cannot be formulated in a measurement model of this form in a time interval $\Delta t_n = [t_k, t_{k+n}]$ of arbitrary length $n$, as required by the preintegration theory.}
\rebuttal{This is because the torque inputs (the measurements) cannot be decoupled from the quadrotor’s orientations (the optimization variables), a fact that was initially noted in VIMO~\cite{nisar2019vimo}.}
\rebuttal{For this reason, the rotational dynamics of the quadrotor are not considered in VIMO, VID, and \methodPrev.}
%
% The preintegration theory requires separating the residual terms that depend on the optimization variables from those that depend on the measurements.
%
% The rotational dynamics of the quadrotor are not considered in \methodPrev, VIMO, and VID as the torque inputs cannot be decoupled from their dependence on the robot's orientation.
%
Instead, these works obtain the evolution of the orientation of the quadrotor from the gyroscope model.
As a result, they introduce inconsistency in the estimation process due to the repeated use of gyroscope measurements, \rebuttal{namely, in the dynamics and IMU residuals}.
Furthermore, their dynamics residuals constrain only the linear dynamics (position and linear velocity) of the quadrotor, while leaving the orientation unconstrained.

In \method\, we address these limitations by representing the rotational dynamics of the quadrotor using a continuous-time formulation.
Specifically, we employ B-splines as the continuous-time function. 
The study in~\cite{cioffi2022continuous} demonstrated that B-splines are well-suited for representing quadrotor dynamics. 
Furthermore, the derivatives of B-splines can be computed efficiently~\cite{sommer2020efficient}, facilitating the use of gradient descent-based optimization methods to optimize the placement of the control points.

\subsubsection{Continuous-time Representation of Rotational Dynamics}\label{sec:ct_quad_dynamics}
We represent the quadrotor body rates $\bm{\omega}$ using a B-spline.
Specifically, we adopt a uniform time representation of the B-spline~\cite{sommer2020efficient}, which allows using a matrix form formulation for sampling.
The B-spline order is denoted by $N$. 
Sampling a point from the B-spline depends only on a local segment defined by N control points.
The control points are placed at the time $t_i = t_0 + i \cdot \Delta t$, $i \in [0, \rebuttal{K]}$, where $t_0$ is the time of the first control point, $i$ is the index of the control point, and $\Delta t$ is the constant time spacing between consecutive control points.
For a given time $t$, the uniform time representation is defined as $u(t) = s(t) - i$, where $s(t) = \frac{t - t_0}{\Delta t}$ represents the index of the B-spline segment between control points $i$ and $i+1$.
The control point $i$ is the leftmost control point affecting the sampling at the time $t$.
\rebuttal{For simplicity of the notation, we use $u$ instead of $u(t)$, in some of the equation presented in this section.}
The quadrotor body rate at the uniform time $u$ is expressed as: $\bm{\omega}(u) = [\bm{\omega}_i, \cdots, \bm{\omega}_{i+N-1}] \bm{M}^{N} \bm{u}$.
The matrix $\bm{M}^{N}$ is the blending matrix, which is constant and precomputed offline once the B-spline order $N$ is known.
The vector $\bm{u}$ contains the base coefficients, where the \textit{j}-th entry is equal to $u^j$.
The time derivative are computed as: $\bm{\omega}^{d}(u) = [\bm{\omega}_i, \cdots, \bm{\omega}_{i+N-1}] a^{d} \bm{M}^{N} \bm{u}$, where $a^{d} = \frac{1}{\Delta t^d}$ and $d$ is the derivative order.
Specifically, we sample the quadrotor angular accelerations with $d=1$.
We optimize the B-spline control points to fit the quadrotor rotational dynamics model, c.f. Eq~\ref{eq:drone_model}, using the measurement model \rebuttal{at time $t_k$}:
\begin{equation}\label{eq:bspline_meas_model}
    \rebuttal{\hat{\bm{\tau}}_{k}^{\bfr} + \bm{\tau}_{res_k}^{\bfr} - \bm{n}_{\tau} = \bm{J} \dot{\bm{\omega}}^{\bfr}(u(t_k)) + \bm{\omega}^{\bfr}(u(t_k)) \times \bm{J} \bm{\omega}^{\bfr}(u(t_k)).}
\end{equation}
\rebuttal{Alternatively, one could use a B-spline to represent the quadrotor’s orientations instead of its angular velocities. However, we choose to represent angular velocities due to the faster convergence observed when optimizing the B-spline control points using the measurement model in Eq.~\ref{eq:bspline_meas_model}. An ablation study on the convergence time of the B-spline fitting problem is included in Sec. A of the appendix.}

\subsubsection{Implementation Details}
We use a B-spline of order 5 and $\Delta t $ = 0.01 s.
The length of the B-spline is set to 0.1 s, resulting in a B-spline represented by 10 control points.
\rebuttal{When a new camera frame arrives, new control points are added to the B-spline at the desired sampling rate, starting from the time of the last control point of the previous B-spline up to the timestamp of the newly arrived camera frame. 
To maintain a fixed-length B-spline, the oldest control points are removed.
The new control points are initialized by interpolating the gyroscope measurements at the desired times. 
At this point, the B-spline optimization begins.
Error terms are computed based on Eq.~\ref{eq:bspline_meas_model}, along with their corresponding Jacobians for each available torque measurement. 
In our experiments, torque measurements are sampled at either 200 Hz (Sec.~\ref{sec:Experiments_benchmark_datasets}) or 100 Hz (Sec.~\ref{sec:Experiments_flights_in_continuous_wind}).
Next, the Hessian matrix is computed using its first-order approximation, as proposed in the Levenberg-Marquardt (LM) algorithm~\cite{press1988numerical}. 
The LM optimization is then performed, with a maximum of 10 inner-loop iterations and 100 outer-loop iterations. 
Residuals, Jacobians, and the Hessian matrix are recomputed at each outer-loop iteration. 
The algorithm terminates early if the solution (i.e., the update to the control point values) falls below a predefined threshold (1e-6 in all our experiments).
Finally, angular velocities are sampled from the optimized B-spline at the control command rate and used to compute the dynamics residuals, as described in Sec.~\ref{sec:dynamics_residuals}.}
%
%% Old Version
% New control points are initialized by interpolating gyroscope measurements at the desired time.
%
% Control points that fall outside the desired time window are simply discarded.
%
% Torque measurements and torque residuals are sampled at a rate of 200 [Hz] and used to derive the residual terms for optimizing the B-spline, based on the measurement model presented in Eq.~\ref{eq:bspline_meas_model}.
%
% These residual terms are optimized using a custom Levenberg-Marquardt algorithm~\cite{press1988numerical} specifically designed for this project to meet the computational requirements of running this optimization within our system on resource-constrained platforms.
%
The programming effort to achieve the integration of the proposed continuous-time-based quadrotor dynamics in a VIO system is a key contribution of our work. 
%
% We will release the code as open-source upon acceptance.

\subsection{Estimation Problem Formulation}
\begin{figure}
    \centering
    \includegraphics[width=1.0\linewidth]{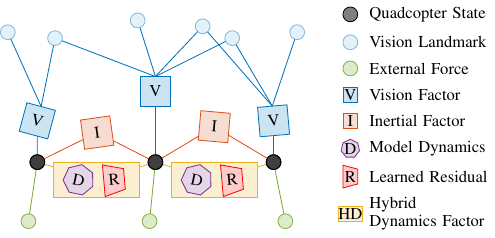}
    \caption{Factor graph representation of \method\ with visual, inertial, and 6-DoF hybrid dynamics factors.}
    \label{fig:factor_graph}
    \vspace*{-18pt}
\end{figure}
We implement our hybrid drone dynamics in a sliding-window optimization-based VIO system.
An overview of the proposed optimization-based VIO with hybrid drone dynamics, using a factor graph representation, is shown in Fig.~\ref{fig:factor_graph}.
The sliding window contains the most recent $L$ keyframes and $K$ drone states. 
We set $L$=10 and $K$=5. 
The optimization variables are defined as: $\mathcal{X} = \{ \mathcal{L}, \mathcal{X}_\mathcal{L}, \mathcal{X}_{\bfr} \}$, where $\mathcal{L}$ consists of the position of the 3D landmarks visible in the sliding window, $\mathcal{X}_\mathcal{L}$ represents the poses of the keyframes: $\mathcal{X}_\mathcal{L} = [\bm{\zeta}_1, \cdots, \bm{\zeta}_l], \space l \in [1,L]$, and $\mathcal{X}_{\bfr}$ the poses of the drone: $\mathcal{X}_{\bfr} = [\bm{x}_1, \cdots, \bm{x}_k], \space k \in [1,K]$.
The pose of the $l^\text{th}$ keyframe is $\bm{\zeta}_l = [\mathbf{p}_{\bfr_l}^{\wfr}, \mathbf{q}_{\bfr_l}^{\wfr}]$, and the state of the $k^\text{th}$ drone is $\bm{x}_{k} = [\bm{p}_{\bfr_k}^{\wfr}, \bm{q}_{\bfr_k}^{\wfr}, \bm{v}_{\bfr_k}^{\wfr}, \bm{b}_{a_k}, \bm{b}_{g_k}, \bm{f}^{\bfr}_{e_k}].$
\rebuttal{Keyframes are selected based on heuristics that consider the magnitude of the estimated motion and the number of visual keypoints being tracked. We adopt the keyframe selection strategy proposed in~\cite{Forster17troSVO}. Visual keypoints are detected only at keyframes.}
The visual-inertial-dynamics estimation problem is formulated as a joint nonlinear optimization that solves for the maximum a posteriori estimate of $\mathcal{X}$. 
The cost function to minimize is:
\begin{align}\label{eq:cost_function}
    \mathcal{L}^{\method} = & \sum_{h = 0}^{L+K-1} \sum_{j \in \mathcal{J}_h} \norm{ \bm{e}_{v}^{j,h} }_{\bm{W}_{\bm{v}}^{j,h}}^{2} + \sum_{k = 0}^{K-1} \norm{ \bm{e}_{i}^{k} }_{\bm{W}_{i}^{k}}^{2} + \nonumber \\ & \sum_{k = 0}^{K-1} \norm{ \bm{e}_{d}^{k} }_{\bm{W}_{d}^{k}}^{2} + \norm{ \bm{e}_{m} }^{2}.
\end{align}
The cost function in Eq.~\ref{eq:cost_function} consists of the visual residuals $\bm{e}_{v}$, inertial residuals $\bm{e}_{i}$, dynamics residuals $\bm{e}_{d}$, and marginalization residuals $\mathbf{e}_{m}$.
All residuals are weighted according to their measurement noise.
The visual residuals are defined as $\bm{e}_{v}^{j,h} = \mathbf{z}^{j,h} - h(\mathbf{l}_{j}^{\wfr})$, representing the re-projection error of the landmark $\mathbf{l}_{j}^{\wfr} \in \mathcal{J}_h$, where $\mathcal{J}_h$ is the set of all the landmarks visible from the frame $h$. 
The function $h(\cdot)$ denotes the camera projection model, and $\mathbf{z}^{j,h}$ represents the corresponding 2D image measurement.
For further details, we refer the reader to~\cite{leutenegger2015keyframe}.
The inertial residuals $\bm{e}_{i}$ are computed using the IMU preintegration algorithm described in~\cite{forster2016manifold}.
The dynamics residuals are detailed in Sec.~\ref{sec:dynamics_residuals}.
The error term $\bm{e}_{m}$ represents prior information obtained from marginalization. 
The marginalization factor encodes information about quantities that fall outside the current sliding window.
We follow the marginalization strategy proposed in~\cite{leutenegger2015keyframe}. 
This approach distinguishes between variables to marginalize (included in the derivation of the marginalization residual) and variables to drop.
\rebuttal{The poses of keyframes and 3D points that are connected to keyframes still included in the optimization window are marginalized.}
Instead, the variables to drop are those not connected to keyframes, such as 3D points visible only from frames outside the sliding window that are not selected as keyframes. 
Dropping these variables, rather than marginalizing them, preserves the sparsity of the Jacobian in Eq.~\ref{eq:cost_function}.
Our implementation of the sliding-window optimization is based on~\cite{leutenegger2015keyframe}. 

We integrate this VIO backend with the visual frontend introduced in~\cite{Forster17troSVO}. 
This decision is motivated by the robustness demonstrated by~\cite{Forster17troSVO}, attributed to its semi-direct approach to visual feature tracking and its low computational requirements. 
This aspect makes it particularly well-suited for VIO applications onboard flying vehicles.
The code for this VIO pipeline is publicly available as open-source\footnote{\url{https://github.com/uzh-rpg/rpg_svo_pro_open}}.
We incorporate the proposed 6-DoF hybrid-dynamics model, the previous 3-DoF hybrid-dynamics model~\cite{cioffi2023hdvio}, as well as the baselines VIMO~\cite{nisar2019vimo} and VID~\cite{ding2021vid} into this VIO system.

\subsection{Dynamics Residuals}\label{sec:dynamics_residuals}
%
% We define the collective thrust measurement model as: $\hat{\bm{f}}^{\bfr}_{k} = \bm{f}^{\bfr}_{t_k} + \bm{f}^{\bfr}_{res_k}  + \bm{n}_{f_{t}}$.
% %
% In addition to the residual force $\bm{f}^{\bfr}_{\text{res}_k}$, we also consider a zero-mean gaussian noise $\bm{n}_{f_{t}} \thicksim \mathcal{N}(0, \bm{\sigma}_{f_{t}}^2)$ to account for uncertainty in the force direction.
% %
% The torque measurement model is: $\hat{\bm{\tau}}^{\bfr}_{k} = \bm{\tau}^{\bfr}_{k} + \bm{\tau}^{\bfr}_{res_k}$.
% %
% The torque measurement $\hat{\bm{\tau}}^{\bfr}_{k}$ is used to optimize the B-spline representing the quadrotor body rates as described in Sec.~\ref{sec:ct_quad_dynamics}.
% %
% The body rate measurement model is: $\hat{\omega}^{\bfr}_k = \omega^{\bfr}_k + \bm{n}_{\omega}$, where $\omega^{\bfr}_k$ is sampled from the B-spline and $\bm{n}_{\omega}$ is a noise value that accounts for uncertainty in the B-spline fitting process.
%
Given two consecutive states at times $t_k$ and $t_{k+1}$, the dynamics motion constraint is:
\begin{align}\label{eq:dynamic_motion_constraint}
    \bm{e}_{d}^{k} = \begin{bmatrix} \bm{\alpha}^{\bfr_k}_{\bfr_{k+1}} - \hat{\bm{\alpha}}^{\bfr_k}_{\bfr_{k+1}} \\ 
    \bm{\beta}^{\bfr_k}_{\bfr_{k+1}} - \hat{\bm{\beta}}^{\bfr_k}_{\bfr_{k+1}} \\
    \bm{\gamma}^{\bfr_k}_{\bfr_{k+1}} - \hat{\bm{\gamma}}^{\bfr_k}_{\bfr_{k+1}} \\
    \bm{f}_{e_{k+1}}^{\bfr} - \hat{\bm{f}}_{e_{k+1}}^{\bfr}\end{bmatrix},
    \bm{W}^{k}_d = \begin{bmatrix} ^{\bm{P}}\bm{W}^{k}_d & \bm{0} \\ \bm{0} & ^{\bm{F}}\bm{W}^{k}_d \end{bmatrix}.
\end{align}
The quantities $\bm{\alpha}^{\bfr_k}_{\bfr_{k+1}}$, $\bm{\beta}^{\bfr_k}_{\bfr_{k+1}}$, and $\bm{\gamma}^{\bfr_k}_{\bfr_{k+1}}$ are the position, velocity, and orientation change in the time interval $[t_k, t_{k+1}]$:
\begin{align}
    \bm{\alpha}_{\bfr_{k+1}}^{\bfr_{k}} &= \bm{R}_{\wfr}^{\bfr_k} (\bm{p}_{\bfr_{k+1}}^{\wfr} - \bm{p}_{\bfr_{k}}^{\wfr} - \bm{v}^{\wfr}_{\bfr_{k}} \Delta t_{k} - \frac{1}{2} \bm{g}^{\wfr} \Delta t_{k}^{2}) \nonumber \\
    &- \frac{1}{2} \bm{f}^{\bfr}_{e_{k}} \Delta t_{k}^2 \nonumber \\
    \bm{\beta}_{\bfr_{k+1}}^{\bfr_{k}} &= \bm{R}_{\wfr}^{\bfr_k} (\bm{v}_{\bfr_{k+1}}^{\wfr} - \bm{v}_{\bfr_{k}}^{\wfr} - \bm{g}^{\wfr} \Delta t_{k}) - \bm{f}^{\bfr}_{e_{k}} \Delta t_{k} \nonumber \\
    \bm{\gamma}^{\bfr_k}_{\bfr_{k+1}} &= \bm{q}^{\bfr_k}_{\bm{W}} \otimes \bm{q}_{\bfr_{k+1}}^{\bm{W}}.
\end{align}
The quantities $\hat{\bm{\alpha}}^{\bfr_k}_{\bfr_{k+1}}$, $\hat{\bm{\beta}}^{\bfr_k}_{\bfr_{k+1}}$, $\hat{\bm{\gamma}}^{\bfr_k}_{\bfr_{k+1}}$ are the preintegrated position, velocity, and orientation.
We calculate them in the discrete time using Euler numerical integration over the timestep $\delta t$:
\begin{align}
    \hat{\bm{\alpha}}_{i+1}^{\bfr_{k}} & = \hat{\bm{\alpha}}_{i}^{\bfr_{k}} + \hat{\bm{\beta}}_{i}^{\bfr_{k}} \delta t + \frac{1}{2}\bm{R}(\hat{\bm{\gamma}}_{i}^{\bfr_{k}})(\hat{\bm{f}^{\bfr}_{i}}+\rebuttal{\bm{f}^{\bfr}_{res_k}}) \delta t^2 \nonumber \\
    \hat{\bm{\beta}}_{i+1}^{\bfr_{k}} & = \hat{\bm{\beta}}_{i}^{\bfr_{k}} + \bm{R}(\hat{\bm{\gamma}}_{i}^{\bfr_{k}})(\hat{\bm{f}^{\bfr}_{i}}+\rebuttal{\bm{f}^{\bfr}_{res_k}}) \delta t \nonumber \\
    \hat{\bm{\gamma}}_{i+1}^{\bfr_{k}} & = \hat{\bm{\gamma}}_{i}^{\bfr_{k}} \otimes \begin{bmatrix} 1 \\ \frac{1}{2}\hat{\bm{\omega}}^{\bfr}_{i}\delta t \end{bmatrix},
    \label{eq:preintegration_terms_recursive_formulation}
\end{align}
where the initial conditions are $\hat{\bm{\alpha}}_{\bfr_{k}}^{\bfr_{k}} = \hat{\bm{\beta}}_{\bfr_{k}}^{\bfr_{k}}$ = 0 and $\hat{\bm{\gamma}}_{\bfr_{k}}^{\bfr_{k}}$ equal to the identity quaternion. The quantity $\bm{R}(\hat{\bm{\gamma}}_{i}^{\bfr_{k}})$ is the rotation matrix representation of $\hat{\bm{\gamma}}_{i}^{\bfr_{k}}$.
\rebuttal{The force residual $\bm{f}^{\bfr}_{res_k}$ is constant in the integration interval.}
%
% We run the propagation algorithm at 100 [Hz], which is the sampling frequency of the dynamics measurements.
\rebuttal{In our experiments, the dynamics measurements are sampled at either 200 Hz or 100 Hz, which corresponds to a timestep $\delta t$ of 5 ms and 10 ms, respectively.}
\rebuttal{The derivation of the covariance of the preintegrated terms is based on the error-state system~\cite{sola2017quaternion} that describes the evolution of the error state $\delta \bm{z} = [\delta \bm{\alpha}, \delta \bm{\beta}, \delta \bm{\theta}]^{\top}$ and noise $\bm{n} = [\bm{n}_{f_t}, \bm{n}_{\omega}]^{\top}$, where $\delta \bm{\theta}$ is a 3-D perturbation around $\hat{\bm{\gamma}}^{\bfr_{k}}$. The discrete-time evolution of the covariance matrix is obtained by linearizing the error-state system in the timestep $\delta t$. The weight assigned to the residual, $^{\bm{P}}\bm{W}^{k}_d$, is the inverse of this covariance matrix.}
%
% The weight assigned to the residual, $^{\bm{P}}\bm{W}^{k}_d$ is the inverse of the covariance matrix derived by linearizing the error $\delta \bm{z} = [\delta \bm{\alpha}, \delta \bm{\beta}, \delta \bm{\theta}]^{\top}$, and noise, $\bm{n} = [\bm{n}_{f_t}, \bm{n}_{\omega}]^{\top}$ in $\delta t$. 
%
The quantity $\hat{\bm{f}}_{e_{k+1}}^{\bfr}$ is the external force preintegrated term.
Following the derivation proposed in~\cite{ding2021vid}, we compute this term as the mean difference between the acceleration measurements and the thrust measurements in the time window $[t_k, t_{k+1}]$: $\hat{\bm{f}}_{e_{i+1}}^{\bfr} = \hat{\bm{f}}_{e_{i}}^{\bfr} + \bm{R}(\bm{\gamma}^{\bfr_k}_i) (\hat{\bm{a}}^{\bfr_i} - \bm{b}_{a_k} - \hat{\bm{f}}^{\bfr}_{i} - \rebuttal{\bm{f}^{\bfr}_{res_k}})$, with $\hat{\bm{f}}^{\bfr}_{e_i} = 0$.
The weight $^{\bm{F}}\bm{W}^{k}_d$ is obtained using the same covariance propagation schema as described above with noise $\bm{n} = [\bm{n}_{a} - \bm{n}_{f_t}, \bm{n}_{\bm{b}_a}]^{\top}$.
This preintegrated term depends on the accelerometer bias. 
To avoid repropagating this term each time the accelerometer bias estimate changes, we adopt the strategy proposed in~\cite{forster2016manifold}. 
Specifically, the preintegration term is corrected by its first-order approximation with respect to the change in the accelerometer bias.

\subsection{Learning Residual Dynamics}
The dynamics residual term described above relies on accurately estimating the forces acting on the vehicle.
In previous works, modeling aerodynamic effects—such as drag forces—requires knowledge of the vehicle's linear velocity, which is not directly measured but instead forms part of the state to be estimated. 
As a result, simply employing a state-of-the-art quadcopter dynamics model is not feasible.

In our approach, we have access to rotor speeds and IMU measurements, as these quantities are directly measured.
Our goal is to estimate residual forces $\bm{f}^{\bfr}_{res}$ and torques $\bm{\tau}^{\bfr}_{res}$, which account for aerodynamic effects and model mismatches, including systematic noise, between the commanded or measured thrust and torque and the actual force acting on the robot in the absence of external disturbances.
To estimate the residual forces, we propose two temporal convolutional networks (TCN).
TCNs have been shown to be as effective as recurrent networks in modeling temporal sequences~\cite{bai2018empirical}, while requiring less computation.
The first TCN is used to predict the residual thrust, and the second TCN is used to predict the residual torque.
We found empirically that using two TCNs produces more accurate predictions than using a single TCN to predict both the residual thrust and torque.
A TCN architecture consists of four temporal convolutional layers with 64 filters each, followed by three temporal convolutional layers with 128 filters each. 
A final linear layer maps the output to a 3-dimensional vector representing the learned residual thrust or torque. 
The network predicting the residual thrust takes a buffer of collective thrust and gyroscope measurements as input.
The network predicting the residual torque takes a buffer of torques and gyroscope measurements as input.
In both cases, the gyroscope measurements are bias-corrected.
The inputs are sampled at 100 Hz and fed into the TCNs as input buffers of 100 ms length, resulting in 10 thrust or torque and 10 gyroscope measurements per buffer. 
We use the Gaussian Error Linear Unit (GELU) activation function.
During training, we model the bias as a random Gaussian variable with zero mean and a standard deviation of 1e-3. 
At deployment, the current bias estimate is used instead.
Given a buffer of measurements over the time interval $\Delta t_{i,j} = t_{j} - t_i$, the TCNs output the residual thrust $\bm{f}^{\bfr}_{res_i}$ and torque $\bm{\tau}^{\bfr}_{res_i}$.
The residual thrust is added to the thrust inputs $\bm{f}^{\bfr}_{t_k}$ with $k \in [t_i, t_j]$ to compute the forces $\hat{\bm{f}}^{\bfr}_{k}$, which account for aerodynamics and robot miscalibration into account.
Similarly, the residual torque is added to the torque inputs $\bm{\tau}^{\bfr}_{k}$ to compute the torques $\hat{\bm{\tau}}^{\bfr}_{k}$.
The corrected torques are used as measurements in the optimization of the B-spline representing the quadrotor body rates, see Sec.~\ref{sec:ct_quad_dynamics}.
The quadrotor body rates are sampled from the B-spline.
These body rates, as well as the corrected thrusts, are used inside the preintegration framework, see Sec.~\ref{sec:dynamics_residuals}, to derive relative velocity, position, and orientation measurements.
We train the neural network that predicts the residual thrust to minimize the MSE loss:
%
%\mathcal{L}^{HD}_{\bm{f}}(\Delta \bm{\alpha}, & \Delta \hat{\bm{\alpha}}, \Delta \bm{\beta}, \Delta \hat{\bm{\beta}}) \nonumber \\ &
\begin{align}\label{eq:thrust_net_loss}
    \rebuttal{\mathcal{L}^{HD}_{\bm{f}}} = \frac{1}{M} \sum_{m = 1}^{M} (\norm{\bm{\alpha}^{\bfr_j}_{\bfr_i} - \hat{\bm{\alpha}}^{\bfr_j}_{\bfr_i}}^{2} + \norm{\bm{\beta}^{\bfr_j}_{\bfr_i} - \hat{\bm{\beta}}^{\bfr_j}_{\bfr_i}}^{2}).
\end{align}
We train the neural network that predicts the residual torque to minimize the MSE loss:
%
% \mathcal{L}^{HD}_{\bm{\tau}}(\Delta \bm{\gamma}, & \Delta \hat{\bm{\gamma}})
\begin{align}\label{eq:torque_net_loss}
    \rebuttal{\mathcal{L}^{HD}_{\bm{\tau}}} = \frac{1}{M} \sum_{m = 1}^{M} \norm{\bm{\gamma}^{\bfr_j}_{\bfr_i} - \hat{\bm{\gamma}}^{\bfr_j}_{\bfr_i}}^{2}.
\end{align}
where $\bm{\alpha}^{\bfr_j}_{\bfr_i}$, $\bm{\beta}^{\bfr_j}_{\bfr_i}$, and $\bm{\gamma}^{\bfr_j}_{\bfr_i}$ are the ground-truth velocity, position, and orientation changes, and $M$ is the batch size.
To learn the aerodynamic effects and systematic noise in the input measurements, the training data is collected under conditions where no external forces act on the drone. 
Furthermore, our training approach does not require ground-truth force data. 
The training data can be generated using a Structure-from-Motion pipeline~\cite{schoenberger2016sfm, cioffi2022continuous} if a motion-capture system is not available.
The neural networks are trained on a laptop running Ubuntu 20.04 with an Intel Core i9 2.3 GHz CPU and an Nvidia RTX 4000 GPU. 
Training is performed using the Adam optimizer with an initial learning rate of 1e-4.
The inference runs either on the laptop or on an NVIDIA Jetson TX2, which is the computing platform onboard the quadrotor.
The TCN inference runs at $\approx$180 Hz on an NVIDIA Jetson TX2, which exceeds the required \unit[100]{Hz} state-update rate of our controllers for agile flight.

\section{Experiments on Benchmark Datasets}\label{sec:Experiments_benchmark_datasets}
In our experiments, we compare our method against \methodPrev\, VIMO, VID, and the same VIO system without the proposed hybrid-dynamics model (hereafter referred to as VIO). 
Following best practices for evaluating VIO algorithms~\cite{Zhang18iros}, we use the following metrics: translation absolute trajectory error ($\text{ATE}_{\text{T}}$ [m]), rotation absolute trajectory error ($\text{ATE}_{\text{R}}$ [deg]), and relative translation and rotation errors. 
These error metrics are computed after aligning the estimated trajectory using the pose-yaw method~\cite{Zhang18iros}. 
For a detailed description of these metrics, we refer the reader to~\cite{Zhang18iros}.
In addition to trajectory evaluation, we evaluate the accuracy of force estimation by computing the root mean squared error (RMSE) between the ground-truth and predicted forces.

\subsection{NeuroBEM Dataset}\label{sec:NeuroBEMDataset}
\subsubsection*{Experimental Setup}
In this set of experiments, we evaluate the hybrid dynamics model independently of the full VIO pipeline. 
Specifically, we evaluate the accuracy of the predicted external force, $\bm f^{\bfr}$, acting on the quadcopter.
For this evaluation, we utilize the NeuroBEM dataset~\cite{bauersfeld2021neurobem}, which features data from indoor drone flights at speeds of up to \unit[65]{km/h}. 
The dataset includes rotor speeds (from which thrust and torque measurements are derived), IMU measurements, and ground-truth force data.
\rebuttal{We use the provided training sequences to train our learned dynamics model, and the provided testing sequences to evaluate its performance, as well as the performance of the baseline methods. In the testing sequences, 30\% of the trajectories are entirely unseen during training, while the remaining 70\% differ in speed and size.}
We compare the force estimation accuracy of our learned dynamics model to several state-of-the-art baselines:
\begin{itemize}
    \item Quadratic Fit: The model used in VIMO and VID.
    \item BEM: A first-principles model that computes forces and torques acting on a propeller by integrating over infinitesimal area elements of the propeller~\cite{gill2017propeller, bauersfeld2021neurobem}.
    \item PolyFit: A data-driven model that relies on polynomial basis functions to capture drone dynamics~\cite{sun2019quadrotor}.
    \item NeuroBEM: A hybrid model that augments the BEM model with a learning-based component~\cite{bauersfeld2021neurobem}.
\end{itemize}
All baselines, except for the Quadratic Fit model, require the vehicle state as input, including linear and angular velocities.
\rebuttal{Notably, our method only requires thrust, torque, and IMU measurements.}
\begin{table}[t]
\centering
\caption{Comparison in terms of RMSE of the force, $F$, and torque, $M$, estimates on the test set of the NeurBEM dataset. Polyfit, NeuroBEM, and \method\ are data-driven methods. Quadratic fit and BEM are first-principles methods. Our method performs remarkably well given it has no information about the velocity or orientation of the vehicle and only falls short of the NeuroBEM method which has access to the ground-truth vehicle state. The values for baseline methods are taken from~\cite{bauersfeld2021neurobem}. In bold are the best values, and in underlined are the second-best values.}
\label{tab:neurobem}
\vspace*{4pt}
\setlength{\tabcolsep}{3pt}
\begin{tabularx}{1\linewidth}{ll|CCCCCC}
\toprule
Model & Inputs & $F_\text{xy}$ [N] & $F_\text{z}$ [N] & $M_\text{xy}$ [Nm] & $M_\text{z}$ [Nm] & $F$ [N] & $M$ [Nn] \\
\midrule
Quadratic Fit & thrust & 1.536  & 1.381  & 0.104 & 0.033 & 1.486 & 0.087 \\
BEM~\cite{bauersfeld2021neurobem} & full state & 0.803 & 1.265 & 0.090 & 0.017 & 0.982 & 0.074 \\
PolyFit~\cite{sun2019quadrotor} & full state & 0.453 & 0.832 & \underline{0.027 }& 0.008 & 0.606 & \underline{0.022} \\
NeuroBEM~\cite{bauersfeld2021neurobem} & full state & \textbf{0.204} & \textbf{0.504} & \textbf{0.014} & \textbf{0.004}  & \textbf{0.335}  & \textbf{0.012} \\
\midrule
\ourMethod & dynamics+gyro  & \underline{0.402} & \underline{0.672} & \textbf{0.014} & \underline{0.006} & \underline{0.491} & \textbf{0.012}\\
% & gyro  & & & & & & \\
% \multirow[c]{2}[-1pt]{\centering \ourMethod} & \multirow[c]{2}[-1pt]{\centering thrust + torque \newline gyro} & 
\bottomrule
\end{tabularx}
\end{table}
\begin{figure}
\centering
\includegraphics[width=1.0\linewidth]{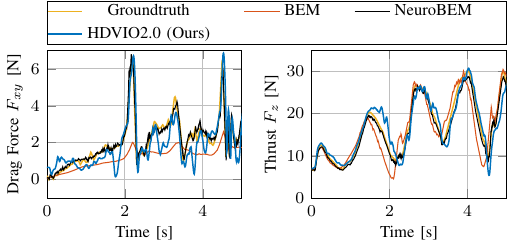}
\caption{This figure illustrates the results shown in Tab.~\ref{tab:neurobem} by exemplarily showing the force estimates on a very fast trajectory from the NeuroBEM dataset. Our \method\ clearly outperforms the state-of-the-art first-principle model BEM and is able to model aerodynamic effects accurately on short timescales.}
    \label{fig:neurobem_plot}
    \vspace*{-12pt}
\end{figure}

\subsubsection*{Evaluation}
The results are summarized in Tab.~\ref{tab:neurobem}. 
The NeuroBEM method, which has access to the full robot state, achieves the best performance. 
However, our \method\ outperforms the BEM model in terms of residual thrust estimates by a factor of three and the PolyFit model by a factor of two. 
This is further illustrated in Fig.~\ref{fig:neurobem_plot}, which shows the ground-truth forces alongside the forces estimated by NeuroBEM, BEM, and our \method\ during the first five seconds of a high-speed flight. 
In this flight, the quadrotor accelerates to \unit[15]{m/s} while following a lemniscate track. 
The performance of our \method\ is remarkable as it achieves this performance without access to ground-truth state information, such as the vehicle's linear or angular velocity, which the baselines require.
Moreover, our \method\ achieves very similar accuracy as NeuroBEM in estimating torques. 
From this experiment, we conclude that our learning-based component effectively captures the aerodynamic forces acting on the quadrotor, demonstrating its suitability for integration into a VIO pipeline. 
Additionally, the ability to estimate aerodynamic forces with such precision using only a history of thrust, torque and gyroscope measurements is an interesting and significant finding in itself.

\subsection{Blackbird Dataset}\label{sec:BlackbirdDataset}
\subsubsection*{Experimental Setup}
In this set of experiments, we evaluate our system and the baseline methods on the Blackbird dataset~\cite{antonini2020blackbird}. 
The dataset provides rotor speed measurements, which we use to compute mass-normalized collective thrust measurements, recorded onboard a quadrotor flying within a motion-capture system. 
Additionally, the dataset includes IMU measurements and, in some sequences, photorealistic images of synthetic scenes.
The Blackbird dataset comprises 18 diverse trajectories with speeds ranging from \unit[0.5]{m/s} to \unit[9.0]{m/s}. 
Since the dataset does not feature external disturbances, we focus on evaluating the accuracy of pose estimates. 
Among the trajectories with available camera images, we select six representative sequences for \rebuttal{testing}: \textit{Bent Dice}, \textit{Clover}, \textit{Egg}, \textit{Mouse}, \textit{Star}, and \textit{Winter}. 
The remaining \unit[80]{\%} of the trajectories, corresponding to approximately \unit[2]{hrs} of flight data, are used for training, and \unit[20]{\%} are reserved for validation of our neural networks.
To assess the generalization capability of our system, we also train our network on a reduced training dataset containing only trajectories with speeds up to \unit[2]{m/s}.
This allows us to evaluate the performance of our method on higher-speed trajectories beyond those seen during training.
\begin{table*}[t!]
\caption{
Evaluation of the trajectory estimates in the Blackbird dataset. \method$^*$ (ours) is trained on a reduced training set, with speeds up to \unit[2]{m/s} to evaluate generalization performance. In bold are the best values, and in underlined are the second-best values. Our method, either using the full training dataset or the reduced one, outperforms the baselines in all the sequences.}
\vspace{3pt}
\label{tab:blackbird}
\setlength{\tabcolsep}{8pt}
\begin{tabularx}{\textwidth}{P{1.2cm}P{0.8cm}|C|C|C|C|C|C}
\toprule 
\multirow[c]{2}{=}[-6pt]{\centering Trajectory\newline Name} & \multirow[c]{2}{=}[-6pt]{\centering $v_{\max}$ \newline [\unit{m/s}]} & \multicolumn{6}{c}{Evaluation Metric: $\text{ATE}_\text{T}$ [\unit{m}] / $\text{ATE}_\text{R}$ [\unit{deg}]}  \\[4pt]
& & VIO & VIMO & VID & \methodPrev & \method\ (Ours) & \method$^*$ (Ours)\\ [4pt]
 \midrule
 Bent Dice & 3 & 0.20 / 1.78 & 0.31 / 1.53 & 0.25 / 1.18 & 0.21 / 1.53 & \underline{0.18} / \textbf{0.84} & \textbf{0.16} / \underline{0.90}\\
 Clover & 5 & 0.90 / 3.52 & 0.88 / 3.66 & 0.83 / 2.48 & 0.60 / 2.08 & \underline{0.49} / \underline{1.99} & \textbf{0.48} / \textbf{1.93}\\
 Egg & 5 & 1.07 / 1.54 & 0.75 / 1.34 & 0.81 / 1.61 & \underline{0.59} / 1.21 & \textbf{0.56} / \textbf{1.07} & \textbf{0.56} / \underline{1.20}\\
 Egg & 6 & 1.40 / 2.35 & 0.98 / 4.89 & 1.10 / 2.42 & 0.83 / \underline{1.62} & \underline{0.69} / \textbf{1.60} & \textbf{0.58} / 1.86\\
 Egg & 8 & 1.79 / 4.55 & 1.57 / 3.69 & 1.47 / 4.84 &  \underline{1.06} / 2.89 & \textbf{0.77} / \underline{2.61} & 1.44 / \textbf{2.51}\\
 Mouse & 5 & 1.10 / 4.54 & 0.76 / 2.14 & 0.54 / 2.10 & 0.36 / 1.40 & \textbf{0.22} / \textbf{0.99} & \underline{0.34} / \underline{1.01}\\
 Star & 1 & 0.17 / 0.78 & 0.18 / 1.05 & 0.18 / 0.54 & 0.16 / \underline{0.58} & \textbf{0.09} / 0.74 & \underline{0.10} / \textbf{0.49}\\
 Star & 3 & 0.62 / 3.50 & 0.43 / 1.38 & 0.50 / 2.93 & 0.38 / 1.40 & \textbf{0.19} / \textbf{0.93} & \underline{0.27} / \underline{0.96}\\
 Winter & 4 & 0.97 / 2.92 & 0.69 / 2.46 & 0.66 / 2.05 & 0.57 / \underline{1.54} & \textbf{0.12} / \textbf{0.78} & \underline{0.51} / 2.02\\
\bottomrule
\end{tabularx}
\end{table*}

\begin{figure}
\centering
\includegraphics[width=1.0\linewidth]{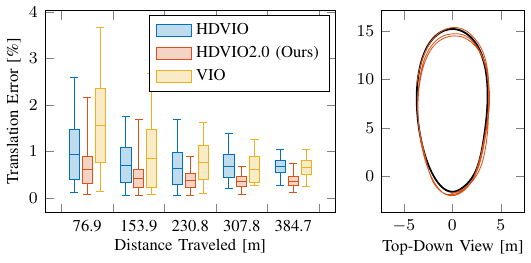}
\caption{The performance of VIO, \methodPrev\ and \method\ is compared on the \emph{Egg} \unit[8]{m/s} trajectory. In the right plot, the ground truth is depicted in black and the trajectory estimated by \method\ in red.}
    \label{fig:blackbird_plot}
    \vspace*{-12pt}
\end{figure}
\begin{table}[!]
    \centering
    \setlength{\tabcolsep}{3pt}
    \caption{Ablation study of the effect of learning residuals (LR) on the trajectory estimates for the Blackbird dataset. In bold are the best values.}
    \begin{tabularx}{1\linewidth}{lC|CC|CC}
    \toprule
         Traj. & $v_{\max}$ [\unit{m/s}] & \multicolumn{2}{c|}{$\text{ATE}_\text{T}$ [\unit{m}]}  & \multicolumn{2}{c}{$\text{ATE}_\text{R}$ [\unit{deg}]} \\
         & & w/ LR & w/o LR & w/ LR & w/o LR \\[2pt]
         \midrule
         Clover & 5 & \bf 0.49 & 0.71 & \bf 1.99 & 2.24 \\
         Egg & 6 & \bf 0.69 & 0.92 & \bf 1.60 & 1.84 \\
         Mouse & 5 & \bf 0.22 & 0.43 & \bf 0.99 & 1.86 \\
         \bottomrule
    \end{tabularx}
    \label{tab:blackbird_ablation}
\end{table}

\subsubsection*{Evaluation}
We present the $\text{ATE}_{\text{T}}$ and $\text{ATE}_{\text{R}}$ results for the \rebuttal{testing} sequences in Table~\ref{tab:blackbird}.
Our approach consistently outperforms VIO baseline, VIMO, VID, and \methodPrev\ demonstrating the effectiveness of our 6-DoF hybrid drone model.
The performance improvement becomes more pronounced at higher speeds, achieving an improvement of the $\text{ATE}_{\text{T}}$ of \unit[57]{\%} and \unit[27]{\%} compared to VIO and \methodPrev\, respectively, at the maximum velocity of \unit[8]{m/s}.
This result is explained by the fact that our method includes the learned drag forces as measurements in the dynamics motion constraint.
In Fig.~\ref{fig:blackbird_plot}, we provide the relative translation error alongside a top-down view of the trajectories estimated by \method\ and the ground truth.
Notably, our system continues to outperform the baselines across almost all sequences, even when the networks are trained on the reduced dataset (containing trajectories with speeds up to \unit[2]{m/s}), as shown in the last column of Table~\ref{tab:blackbird}.
\camera{This result highlights that \method\ is capable of generalizing to velocities up to 4x higher than those present in the training data.}
The fact that the accuracy of \method\ and \method$^*$ is higher than \methodPrev\ in most of the sequences highlights the benefit of incorporating the rotational dynamics in the dynamics residuals.
In Table~\ref{tab:blackbird_ablation}, we show an ablation study to evaluate the effect of learning residual thrusts and torques.
The results on the faster sequences of the Blackbird dataset show that including learned residuals in the drone dynamics improves the trajectory estimates.

\subsection{VID Dataset}\label{sec:VIDDataset}
\subsubsection*{Experimental Setup}
In this set of experiments, we evaluate the ability of our system to estimate external forces acting on the quadrotor and test the performance of our learned component in scenarios where ground-truth data from a motion capture system is unavailable. 
For this purpose, we use the VID dataset~\cite{ding2021vid}, which includes visual, inertial, and actuation inputs, as well as ground-truth force measurements.
The dataset contains data recorded onboard a quadrotor flying both indoors, in an office room equipped with a motion-capture system, and outdoors, in a parking area. 
We use the provided rotor speed measurements to compute the thrust and torque values. 
The indoor sequences, which include portions with ground-truth force data, are used to evaluate our system's capability to estimate external forces. 
The outdoor sequences are used to validate our learned module when ground-truth training data for position, velocity, and orientation is obtained from an offline visual-inertial SLAM system~\cite{cioffi2022continuous} rather than a motion-capture system.
Since the outdoor sequences do not include ground-truth force measurements, we focus on the estimation of the drone poses. 
Since the quadrotor mass differs between the indoor and outdoor sequences, we train different neural networks for the indoor drone configuration and the outdoor configuration.
For the indoor configuration, we train our neural networks using sequences without external perturbations and ground truth from the motion capture system. 
These sequences consist of hover, circle, and figure 8 trajectories, amounting to only \unit[6]{min} of flight data. 
Of this data, $80\%$ is used for training and $20\%$ for validation.
For the outdoor configuration, the sequences include circle, figure 8, and rectangle trajectories, and the ground truth is obtained by the offline visual-inertial SLAM system. 
We use one figure 8 and one rectangle trajectory for testing. 
The remaining data, totaling \unit[11]{min} of flight time, is split into $80\%$ for training and $20\%$ for validation.
\begin{table}[!]
    \centering
    \setlength{\tabcolsep}{3pt}
    \caption{\rebuttal{RMSE of the external force estimates. The external force originates either from a pulling rope (\textit{sequence 17} of the VID dataset) or from an external load (\textit{sequence 16} of the VID dataset).} \method\ drastically improves the force estimation along the z axes. In bold are the best values.}
    \begin{tabularx}{1\linewidth}{l|CC|C}
    \toprule
         Method & \multicolumn{2}{c|}{Pulling rope}  & External Load\\
         & $F_\text{z}$ [N] & $F$ [N] & $F_\text{z}$ [N]  \\[2pt]
         \midrule
         VIMO & 1.73 & 1.08 & 0.81 \\
         VID & 1.96 & 1.12 & 0.45 \\
         \methodPrev & 0.55 & 0.65 & \textbf{0.34} \\
         \ourMethod & \textbf{0.39} & \textbf{0.59} & \textbf{0.34} \\
         \bottomrule
    \end{tabularx}
    \label{tab:vid_rope_and_load}
\end{table}
\begin{figure}
    \centering
    \includegraphics[width=1.0\linewidth]{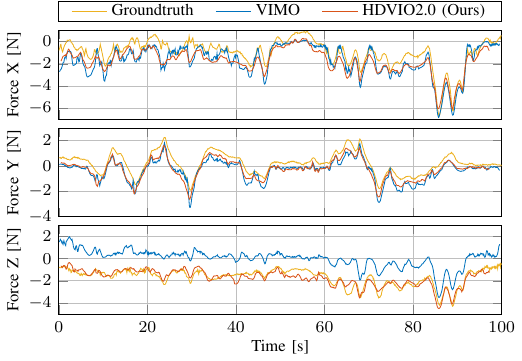}
    \caption{Comparison of the external force estimate in the \textit{sequence 17} of the VID dataset. In this sequence, the drone is attached to an elastic rope. The other end of the rope is attached to a force sensor. \method\ drastically improves the force estimation along the z-axis.}
    \label{fig:viddataset_rope}
    \vspace{-9pt}
\end{figure}
\begin{table}[!]
    \centering
    \setlength{\tabcolsep}{3pt}
    \caption{We demonstrate the performance of \method\ in a setting where the training data for the learning-based component is obtained from a vision-based SLAM system in the outdoor sequences of the VID dataset. The flown trajectories are at low speeds, below \unit[3]{m/s}, allowing all the methods to perform well, with \method\ showing the highest accuracy.
    In bold are the best values.}
    \begin{tabularx}{1\linewidth}{l|CC|CC}
    \toprule
         Method & \multicolumn{2}{c|}{Figure 8}  & \multicolumn{2}{c}{Rectangle}\\
         & $\text{ATE}_\text{T}$ [\unit{m}] & $\text{ATE}_\text{R}$ [\unit{deg}] & $\text{ATE}_\text{T}$ [\unit{m}] & $\text{ATE}_\text{R}$ [\unit{deg}] \\[2pt]
         \midrule
         VIO & 1.89 & 4.02 & 2.09 & 2.01 \\
         VIMO & 1.67 & 3.70 & 1.89 & 1.92 \\
         VID & 1.84 & 3.50 & 1.99 & 2.49 \\
         \methodPrev & 1.48 & 3.70 & 1.72 & 1.68\\
         \ourMethod & \textbf{1.41} & \textbf{0.71} & \textbf{1.69} & \textbf{1.61} \\
         \bottomrule
    \end{tabularx}
    \label{tab:vid_outdoor}
\end{table}

\subsubsection*{Evaluation}

We compare the external force estimates of VIMO, VID, \methodPrev, and \method\ in an indoor sequence where the drone is attached to an elastic rope, with the other end connected to a force sensor, and in a sequence where an unknown external load is attached to the drone, as summarized in Table~\ref{tab:vid_rope_and_load}.
The force estimates are aligned with the motion-capture reference frame using the \textit{posyaw} alignment method~\cite{Zhang18iros}.
\method\ significantly outperforms the baselines, VIO and VID, while the performance increase compared to \methodPrev\ is smaller.
As shown in Fig.~\ref{fig:viddataset_rope}, our method significantly improves force estimation along the z-axis. 
This improvement is attributed to our neural network.
We believe that the proposed network has learned to compensate for a systematic residual error in the thrust inputs, likely caused by inaccuracies in the thrust coefficients used to compute the collective thrust from rotor speed measurements.
In these sequences, the slow vehicle motion and the textured environment simplify the pose estimation problem. 
Consequently, VIO, VIMO, VID, \methodPrev\ and our method achieve similar pose estimation performance, with a $\text{ATE}_{\text{T}}$ of \unit[0.02]{m} and of \unit[0.10]{m}, respectively.

The Table~\ref{tab:vid_outdoor} presents the evaluation of pose estimates in the outdoor sequences. 
Since the trajectories are flown at low speeds below \unit[3]{m/s}, all three methods demonstrate good performance, with \method\ achieving the highest accuracy.
Remarkably, including the rotational dynamics in the estimation process improves the rotation error by $80\%$ compared to the previous system \methodPrev\ in the sequence \textit{Figure 8}.
This experiment highlights that our learning-based dynamics model can be effectively trained without relying on an external motion-capture system.

\section{Flights in Continuous Wind}\label{sec:Experiments_flights_in_continuous_wind}
\begin{figure}
    \centering
    \includegraphics[width=1.0\linewidth]{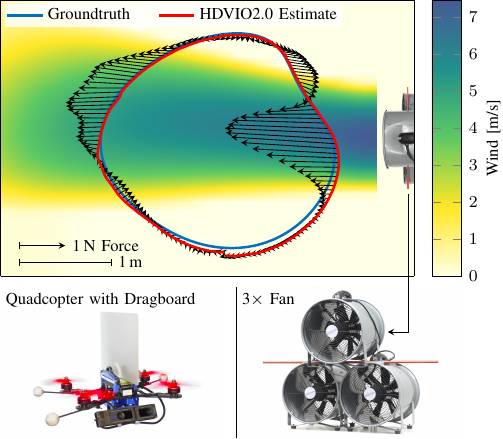}
    \vspace*{6pt}
    \caption{A quadrotor with a dragboard attached is flown on a circular trajectory through a wind field generated with three industrial fans. Our \method\ is used to estimate the position of the drone (shown in red) and the external disturbance force (black arrows) acting on the vehicle. The ground-truth position of the vehicle is shown in blue.}
    \vspace*{-12pt}
    \label{fig:drone_and_wind}
\end{figure}
In these experiments, we demonstrate that \method\ can estimate continuous external disturbances, such as continuous wind, outperforming all the state-of-the-art methods. 
To this end, we fly a quadrotor in a controlled wind field, as illustrated in Fig.\ref{fig:drone_and_wind}. 
Details about the quadrotor platform can be found in~\cite{foehn2022agilicious}. 
The platform is equipped with an onboard Intel RealSense T265\footnote{\url{https://www.intelrealsense.com/wp-content/uploads/2019/09/Intel_RealSense_Tracking_Camera_Datasheet_Rev004_release.pdf}} camera, which provides camera and IMU measurements. 
While the camera includes stereo fisheye sensors, we only use images from the left camera.
Rotor speed measurements are not available on this quadrotor platform. 
Instead, we use the collective thrust commands output by the MPC controller~\cite{foehn2022agilicious} to control the vehicle. 
The quadrotor is flown in a motion-capture system that provides pose data at an update rate of \unit[200]{Hz}.
We conduct experiments using two quadrotor configurations: one with the vehicle in its nominal state and another with a {\unit[22]{cm}$\times$\unit[16]{cm}} dragboard attached. 
The dragboard is mounted such that its normal aligns with the drone’s body y-axis. 
This attachment increases drag in the y-direction by more than a factor of two, making the vehicle significantly more sensitive to crosswinds.

\subsubsection{Wind Generation}
To generate a wind field, we placed three axial fans (Ekström 12 inch, as shown in Fig.~\ref{fig:drone_and_wind}) in an office-like room measuring \unit[8x10]{m}. 
The fans were positioned at a height of \unit[1.6]{m} above the ground and angled slightly inwards to ensure high wind speeds across a virtual tube in front of them. 
Each fan has an advertised air circulation rate of $\unit[1.3]{m^3/s}$ and produces a measured wind speed of up to \unit[8]{m/s} at its front grill.
To quantitatively evaluate the performance of our method, we require ground-truth data for the external wind forces. 
In the following, we describe the procedure used to obtain this ground-truth data.

\subsubsection{Wind Speed Map}
In the first step, we measured the wind produced by our experimental setup. 
Local wind speeds were recorded at 50 points within the wind cone in front of the fans, with a higher sampling density in regions where wind speed changes more rapidly. 
Measurements were taken using a hand-held anemometer (Basetech BS-10AN), and the position of the anemometer was tracked using the motion-capture system. 
To generate the ground-truth wind speed map shown in Fig.~\ref{fig:drone_and_wind}, a smoothing spline was fitted to the recorded data.

\subsubsection{Lift and Drag Coefficients}
We compute the wind force based on the wind speed map. 
The aerodynamic forces acting on a quadrotor are primarily determined by three components: the body/fuselage drag $f_d^\text{fus}$, the induced drag from the propellers $f_{d}^\text{ind}$, and the lift and drag incurred by the flat-plate drag board attached to the top of the quadrotor, $f_l^\text{brd}$ and $f_d^\text{brd}$.
The magnitudes of these forces can be approximated using established aerodynamic models~\cite{bauersfeld2021neurobem, bauersfeld2022rangeestimates, ducard2014modeling}:
\begin{equation}
\begin{aligned}
    f_d^\text{fus} &= 0.5 \, \rho \, A^\text{fus} \, c_d^\text{fus} \, v_\text{rel}^2\; \\
    f_d^\text{ind} &= k \, v_\text{rel}\; \\
    f_{l|d}^\text{brd} &= 0.5 \, \rho \, A^\text{brd} \, c_{l|d}^\text{brd}(\alpha) \, v_\text{rel}^2\;,
\end{aligned}
\label{eq:aero_forces}
\end{equation}
where $\rho$ is the air density, $A$ is a surface area, $v_\text{rel}$ the relative air speed, $\alpha$ is the angle of attack of the dragboard, $k$ is the propeller drag coefficient, and $c_{l|d}$ are the lift and drag coefficients of the fuselage and dragboard. 
The relative air speed is calculated as the norm of the relative velocity, which is the sum of the vehicle's ego motion and the wind velocity.

In this model, the fuselage is approximated as a square prism with an angle-of-attack-independent drag coefficient of $c_d^\text{fus} = 2.0$ \cite{hinsberg2021prismaero}. 
For the flat-plate wing, we employ a simplified high-angle-of-attack model widely used in propeller modeling~\cite{gill2017propeller, ducard2014modeling}, which also aligns well with experimental data for flat-plate wings~\cite{sheldahl1981aerodynamic}:
\begin{align*}
    c_l^\text{brd}(\alpha) = \sin\left( 2\alpha\right)\;, && c_d^\text{brd}(\alpha) = 2 \sin^2(\alpha).
\end{align*}
To validate the predicted forces for the fuselage and drag board described in Eq.~\eqref{eq:aero_forces}, the quadrotor was mounted on a load cell\footnote{\url{https://www.ati-ia.com/products/ft/ft_models.aspx?id=Mini40}}. 
At a wind speed of \unit[7]{m/s}, the measured lift and drag forces were within \unit[10]{\%} of the calculated values.
Additionally, the linear propeller drag coefficient was determined to be \unit[$k=0.145$]{Ns/m}.

\subsubsection{Wind Forces}
Our method distinguishes between aerodynamic effects (such as body drag and induced drag) and external forces. 
To obtain the ground truth for the disturbance caused by the wind, we calculate the forces acting on the quadrotor with the fans turned on and subtract the forces calculated when the fans are turned off.
\begin{table}[!]
    \centering
    \setlength{\tabcolsep}{2pt}
    \caption{Trajectories estimates in drone flights in a wind field with wind gusts up to \unit[25]{km/h}. We use (d) to indicate that a drag board was attached to the drone. In bold are the best values.}
    \begin{tabularx}{1\linewidth}{l|C|C|C|C}
    \toprule
         &  \multicolumn{4}{c}{$\text{ATE}_\text{T}$ [\unit{m}] / $\text{ATE}_\text{R}$ [\unit{deg}]} \\[1pt]
         Method & Circle (d) & Circle & Lemniscate (d) & Lemniscate \\[2pt]
         \midrule
          VIO & 0.07 / 2.02 & 0.06 / 1.21 & 0.38 / 2.39 & 0.27 / 2.44 \\
          VIMO & 0.10 / 1.80 & 0.08 / 1.19 & 0.34 / 2.93 & 0.32 / 1.93 \\
          VID & 0.10 / 2.31 & 0.06 / 1.37 & 0.53 / 2.39 & 0.28 / 2.05 \\
          \methodPrev & 0.07 / 2.06 & 0.06 / 1.17 & 0.30 / 2.81 & 0.20 / 1.84\\
          \ourMethod & \bf 0.05 / 1.02 & \bf 0.04 / 1.02 & \bf 0.21 / 2.34 & \bf 0.14 / 1.53 \\
         \bottomrule
    \end{tabularx}
    \label{tab:flyingroom_ate}
\end{table}
\begin{table}[!]
    \centering
    \setlength{\tabcolsep}{2pt}
    \caption{\rebuttal{RMSE of the external force estimates. The external force originates from continuous wind, with gusts up to \unit[25]{km/h}.} We use (d) to indicate that a dragboard was attached to the drone. In bold are the best values.}
    \begin{tabularx}{1\linewidth}{l|C|C|C|C}
    \toprule
         &  \multicolumn{4}{c}{ $F$ [N]} \\[1pt]
         Method & Circle (d) & Circle & Lemniscate (d) & Lemniscate \\[2pt]
         \midrule
          VIMO & 0.62 & 0.26 & 0.52 & 0.44 \\
          VID & 0.56 & 0.33 & 0.73 & 0.51 \\
          \methodPrev & 0.54 & 0.23 & 0.44 & 0.33\\
          \ourMethod & \bf 0.51 & \bf 0.23 & \bf 0.39 & \bf 0.31 \\
         \bottomrule
    \end{tabularx}
    \label{tab:flyingroom_forces}
\end{table}
\begin{figure*}
    \centering
    \includegraphics[width=1.0\linewidth]{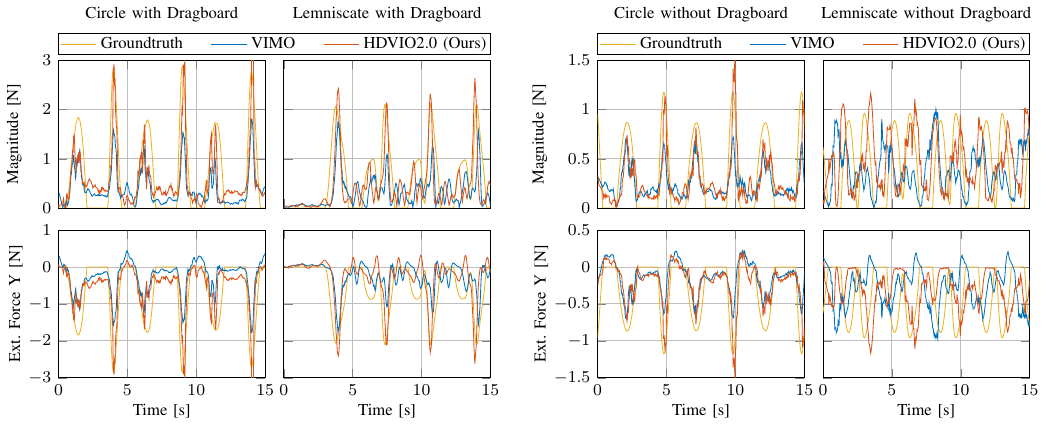}
\caption{Wind disturbance estimates. The magnitude and the y-axis component of the wind force estimated by \method\ and VIMO. Left: drone equipped with a dragboard. Right: standard drone configuration. In all the plots, it is visible that \method\ achieves more accurate force estimates than VIMO.}
    \label{fig:forces_realworld}
    \vspace*{-9pt}
\end{figure*}
\begin{figure}
    \centering
    \includegraphics[width=1.0\linewidth]{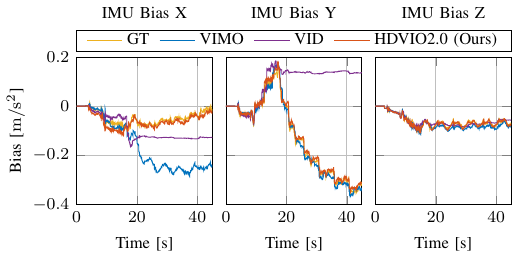}
\caption{Accelerometer bias estimates. The ground-truth (GT) bias is obtained from the VIO system. The estimates of our \method\ match the ground-truth values, while VIMO and VID estimates diverge along the x-axis and the y-axis.}
    \label{fig:biases_realworld}
    \vspace*{-6pt}
\end{figure}

\subsection{Dataset Collection}\label{sec:RealworldExp}
The training data consists of approximately \unit[10]{min} of random trajectories flown without wind. 
We use $80\%$ of this data for training the neural network and the remaining $20\%$ for validation.
We exclusively use \emph{random} trajectories, generated by sampling position data with a Gaussian Process.
This method ensures diverse training data and helps prevent overfitting to specific trajectories. 
The test data is collected with the quadrotor flying in a wind field with gusts reaching up to \unit[25]{km/h}.
The test trajectories include a circle and a lemniscate, both with a maximum speed of \unit[2]{m/s}.
Additionally, we recorded a second dataset that features the same training, validation, and test trajectories, but with the quadrotor equipped with a drag board. 
In this setup, the drag and the external force due to wind gusts are increased, emphasizing the advantage of \method\ over the baselines.

\subsubsection{Evaluation}\label{sec:RealworldExpEval}
We present the estimation of the external force due to wind gusts in Fig.~\ref{fig:forces_realworld} and Table~\ref{tab:flyingroom_forces}.
Since the wind gusts impact the quadrotor along the y-axis of the world reference frame, we display both the y-component of the estimated force and the force norm. 
The external forces are estimated in the quadrotor’s body frame, then aligned to the world frame (which corresponds to the motion-capture reference frame) using the ground-truth orientations. 
This alignment allows for a direct comparison between the estimates of our method and those from the baselines. 
Notably, \method\ is able to accurately predict the wind gusts when the quadrotor enters the wind field. 
This is evident in Fig.~\ref{fig:forces_realworld} from the fact that our method accurately captures the peaks of the wind force.

As noted by the authors~\cite{nisar2019vimo}, the measurement model in VIMO, when dealing with continuous external disturbances, introduces an inconsistency in the estimation of the accelerometer bias, leading to decreased motion estimate accuracy. 
We show in Fig.~\ref{fig:biases_realworld} the accelerometer bias estimated by the VIO algorithm, VIMO, VID, and our method for a sequence in which the quadrotor, equipped with the drag board, flies a circle trajectory.
Although we do not have access to the ground-truth accelerometer bias, we consider the one estimated by the VIO algorithm to be a good approximation of the true value, given that VIO achieves very high performance in this sequence (see Table~\ref{tab:flyingroom_ate}) due to the high number of visual features being tracked. 
The bias estimate of \method\ closely tracks the VIO estimate, while VIMO and VID estimate diverges along the x-axis and y-axis to an incorrect value.
We include the position and orientation absolute trajectory errors in Table~\ref{tab:flyingroom_ate}. 
In all four sequences, the rich-texture environment simplifies the pose estimation problem, resulting in accurate pose estimates for all the algorithms.
However, improvements in the rotation estimates, driven by the inclusion of rotational dynamics in the estimation process, are evident.

\rebuttal{Table~\ref{tab:runtime} reports the runtime of \method\ on two platforms: a laptop running Ubuntu 20.04 with an Intel Core i9 2.3 GHz CPU and an Nvidia RTX 4000 GPU, and an NVIDIA Jetson TX2 (the onboard computer used during flight). 
We break down the runtime into two components corresponding to the frontend and backend threads of \method.
The frontend runtime measures the time required for visual feature tracking and detection. 
The backend runtime is the time from the arrival of a new camera frame to the estimation of its pose, including the setup and solution of the optimization problem defined in Eq.~\ref{eq:cost_function}. 
This includes both the B-spline optimization and neural network inference.
\method\ achieves real-time performance, namely the pose of a camera frame is estimated before the next frame is available (given a 30 FPS camera as in all our experiments) on both the laptop and the NVIDIA Jetson TX2.}
\begin{table}[!]
    \centering
    \setlength{\tabcolsep}{2pt}
    \caption{\rebuttal{Runtime [ms] of \method\ . The backend runtime includes the B-spline optimization and network inference.}}
    \begin{tabularx}{1\linewidth}{C|C|C}
    \toprule
         \method &  \multicolumn{2}{c}{ Runtime [ms]} \\[1pt]
         thread & Laptop & Nvidia Jetson TX2 \\[2pt]
         \midrule
          Frontend & 5.4 & 14.9 \\
          Backend & 21.3 & 31.8\\
         \bottomrule
    \end{tabularx}
    \label{tab:runtime}
\end{table}

\section{Closed-loop Control Flights}\label{sec:closed_loop_control}

\begin{figure}
    \centering
    \includegraphics[width=1.0\linewidth]{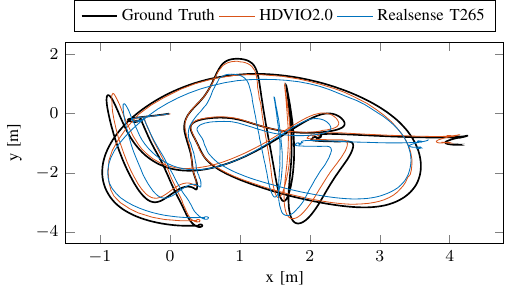}
\caption{Closed-loop experiment where the quadrotor state estimated by \method\ is provided to the flight controller. The figure shows the top view of the trajectory estimated by \method, the Realsense T265, and the ground truth from a motion capture system. Our method outperforms the commercial stereo-based SLAM from the Realsense T265.}
    \label{fig:closed_loop}
    % \vspace*{-6pt}
\end{figure}

In this section, we demonstrate the performance of \method\ running onboard a quadrotor within a closed-loop control system. We use the same drone described in Section~\ref{sec:Experiments_flights_in_continuous_wind} and adopt the controller from~\cite{foehn2022agilicious}.

\method\ provides state estimates at the camera frame rate (30 Hz in our experiments), which is insufficient for high-frequency feedback control that requires estimates at a frequency higher than 100 Hz. To bridge this gap, we integrate \method pose estimates into an extended Kalman filter (EKF). The EKF state includes the quadrotor’s full 6-DoF pose, velocity, and IMU biases. It is propagated using the IMU measurements and updated with the pose estimates from \method. As a result, the EKF provides state estimates at 200 Hz, matching the IMU rate. These high-frequency estimates are then supplied to the controller.

Figure~\ref{fig:closed_loop} compares the trajectory estimated by \method, the onboard Intel RealSense T265, and ground truth. Notably, \method\ outperforms the commercial SLAM system running onboard the T265. Indeed, \method\ achieves an $\text{ATE}_{\text{T}}$ of 0.18 m, while the Intel RealSense T265, which employs a stereo camera configuration, achieves 0.29 m.

A clip of this experiment is included in the supplementary video.

\rebuttal{\section{Discussion}\label{sec:discussion}
A key component of \method\ is the learning-based drone dynamics model, which estimates residual aerodynamic forces using only thrust, torque, and gyroscope measurements, sensors commonly available on most commercial drone platforms. 
This learning-based module outperforms state-of-the-art first-principles quadrotor models and achieves performance comparable to NeuroBEM, despite not having access to the ground-truth quadrotor state as these methods do.

Another important advantage is that it does not necessarily require ground-truth forces for supervised training.
Instead, the training minimizes the error between predicted and observed changes in position, velocity, and orientation, which can be obtained from SLAM~\cite{cioffi2022continuous} or Structure from Motion~\cite{schoenberger2016sfm}, avoiding the need for motion-capture systems. 
Table~\ref{tab:vid_outdoor} shows successful training using only SLAM-derived supervision.
In this way, the data collection is simplified to the extent that even a human pilot can control the drone, as expert pilots typically issue collective thrust commands alongside desired body rates~\cite{pfeiffer2022visual}. 
This simple training data collection mitigates a limitation of our drone dynamics model: while a single dynamics model is tailored to a specific drone, recording new data to train a model for a different drone is straightforward.

\method\ drone dynamics model demonstrates strong generalization capabilities to velocities and trajectories unseen in the training data. 

Generalization to unseen velocities is shown in Table~\ref{tab:blackbird} where \method$^*$ is trained with trajectories that contain speeds only up to \unit[2]{m/s}, compared to \method\ which is trained on speeds up to \unit[9]{m/s}.
When tested on the full range of speeds (up to \unit[8]{m/s}), \method$^*$ achieves comparable results to \method\ and still outperforms the baselines.

Generalization across different trajectory types is shown in various experiments. 
In the NeuroBEM dataset (see Sec.~\ref{sec:NeuroBEMDataset}), 30\% of the test trajectories were entirely unseen during training, while the remaining 70\% differ in speed and size. 
Fig.~\ref{fig:viddataset_rope} shows the force estimates of \method\ on an unseen trajectory from the VID dataset.
In Sec.~\ref{sec:RealworldExp}, the network is exclusively trained on random trajectories. 
These advantages highlight \method\ robustness and versatility in real-world applications.

Furthermore, \method\ demonstrates robustness to VIO failures.
In Sec.~\ref{sec:BlackbirdDataset}, \method\ achieves the largest improvements, with reductions of 57\% and 43\% in translation and rotation error compared to the VIO on the fastest trajectory, Egg 8 m/s (see Table~\ref{tab:blackbird} and Fig~\ref{fig:blackbird_plot}). 
In this scenario, motion blur and fast yaw changes make feature tracking difficult, causing the VIO system to accumulate significant drift. 

In this work, as in the baselines, the model is assumed to remain fixed during a flight. 
Changes in actuation inputs (e.g., hardware degradation) are treated as external forces.
A promising direction for future work would be to train the neural network to estimate these model changes as residual forces, overcoming the challenge of generating suitable training data.

The core idea behind \method\, learning residual dynamics to derive motion constraints for inclusion in a VIO backend, is not specific to quadrotors and can be applied to other robotic platforms.
The feasibility of learning residual dynamics has been demonstrated on diverse robot types, including rigid~\cite{zeng2020tossingbot} and soft~\cite{gao2024sim} robotic arms, and legged robots~\cite{hwangbo2019learning}.
Since VIO backends are robot-agnostic, incorporating motion constraints based on the learned dynamics is generally applicable.
However, specific system design choices, including the selection of the network architecture, may need to be adapted based on the specific dynamics of the target robot.
For instance, while TCNs have proven effective for modeling residual dynamics~\cite{bauersfeld2021neurobem}, motion priors~\cite{cioffi2023learned}, and control commands~\cite{xing2024contrastive} in quadrotors, they may not be optimal for other types of robots.
Assessing their application for different robotic platforms is an interesting direction for future work.}
\section{Conclusion}\label{sec:conclusion}

In this work, we introduce a novel method for modeling 6-DoF quadrotor dynamics in visual-inertial odometry systems. 
Our dynamics model integrates a first-principles quadrotor model with a learning-based component that captures unmodeled effects, such as aerodynamic drag. 
The proposed method addresses the limitations of the state-of-the-art systems, VIMO, VID, and \methodPrev, improving the accuracy of motion estimation and external force estimation.

Our learning-based component demonstrates strong generalization capabilities to trajectories and speeds beyond those present in the training dataset. 
Furthermore, an evaluation of residual force estimation accuracy reveals that our learning-based approach outperforms first-principles models, even those with access to the full state of the quadrotor. 
Controlled experiments in windy conditions further validate our hybrid dynamics model's ability to accurately predict forces acting on the quadrotor due to continuous wind.

\method\ enhances the safety of autonomous drone operations in challenging scenarios, such as high-speed flights and operations in windy environments. 
With the growing integration of drones into everyday applications, these improvements are relevant. 
%
% We believe that our work represents a step forward in ensuring safer and more reliable autonomous flight.
\section*{\rebuttal{Appendix}}\label{sec:appendix}
\subsection{\rebuttal{Representation of Rotational Dynamics}}\label{sec:bspline_repres_appendix}
\rebuttal{In \method\, we represent the quadrotor's angular velocities with a B-spline. Alternatively, it is possible to use a B-spline to represent the quadrotor's orientations.}
\rebuttal{ Our choice to represent angular velocities results from the faster convergence observed in the optimization problem when optimizing the B-spline control points according to the measurement model defined in Eq.~\ref{eq:bspline_meas_model}.}
\rebuttal{To demonstrate this, we conducted an ablation study comparing the convergence time of the B-spline fitting optimization problem when the B-spline represents either the quadrotor’s orientations (\textit{RotBSpl}) or its angular velocities (\textit{VelBSpl}).}
\rebuttal{The results of this study are presented in Table~\ref{tab:convergence_runtime_appendix}.}
\rebuttal{We selected a sequence from the NeuroBEM dataset where the drone follows a random trajectory.}
\rebuttal{This trajectory was split into sequential, non-overlapping segments of durations 0.5, 1, and 2 s.}
\rebuttal{For each segment, we solved the B-spline fitting optimization problem using all available torque inputs as measurements (the NeuroBEM dataset provides control commands at 400 Hz).}
\rebuttal{The control points for \textit{RotBSpl} were initialized via linear interpolation of the quadrotor’s orientations provided by the dataset, while those for \textit{VelBSpl} were initialized using linear interpolation of the gyroscope measurements.}
\rebuttal{We evaluated different B-spline orders and control point spacings. The experiments are run on a laptop running Ubuntu 20.04 and equipped with an Intel Core i9 2.3 GHz CPU.}
\rebuttal{As shown in Table~\ref{tab:convergence_runtime_appendix}, the B-spline representing angular velocities consistently achieved faster convergence than the one representing orientations, regardless of B-spline order, control point spacing, or segment length.}
\begin{table}[!ht]
    \centering
    
    \caption{\rebuttal{Convergence time in [ms] of the optimization problem that optimizes the B-spline control points according to the quadrotor's rotational dynamics model (see Eq.~\ref{eq:bspline_meas_model}).  We compare two B-spline formulations. \textit{RotBSpl}: the control points represent the quadrotor orientation in \textit{SO(3)}. \textit{VelBSpl}: the control points represent the quadrotor angular velocity. \textit{VelBSpl} is the B-spline representation used in HDVIO2.0 because of its fast convergence time. We report the average convergence time computed over all the trajectory segments. In \textbf{bold} the smallest convergence time. -- indicates an infeasible configuration caused by a trajectory segment that is too short relative to the chosen B-spline order and control point spacing.}
} 
    \setlength{\tabcolsep}{2.0pt}
    \renewcommand{\arraystretch}{1.2}
    \begin{tabular}{c|c|cc|cc|cc}
    \toprule
    \textbf{B-spline} & \textbf{Ctrl. Points} & \multicolumn{6}{c}{\textbf{Trajectory Segment Length} [s]} \\[2pt]
    \textbf{Order} & \textbf{Spacing} & \multicolumn{2}{c|}{0.5 [s]} & \multicolumn{2}{c|}{1.0 [s]} & \multicolumn{2}{c}{2.0 [s]}\\
    & [ms] & \textit{RotBSpl} & \textit{VelBSpl} & \textit{RotBSpl} & \textit{VelBSpl} & \textit{RotBSpl} & \textit{VelBSpl}\\
    \midrule
     & 100 & -- & -- & 500.8 & \textbf{12.4} & 1498.2 & \textbf{40.3}\\
     & 50 & 272.3 & \textbf{7.6} & 707.5 & \textbf{21.4} & 1720.9 & \textbf{34.9}\\
     5 & 20 & 431.6 & \textbf{7.9} & 914.0 & \textbf{17.4} & 1881.3 & \textbf{29.3}\\
     & 10 & 536.3 & \textbf{3.6} & 1082.0 & \textbf{9.3} & 2243.5 & \textbf{16.2}\\
     & 5 & 702.6 & \textbf{8.3} & 1384.8 & \textbf{19.6} & 2795.7 & \textbf{28.8}\\
    \midrule
     & 100 & -- & -- & 509.3 & \textbf{6.2} & 1676.4 & \textbf{16.9}\\
     & 50 & 278.8 & \textbf{3.4} & 805.2 & \textbf{12.1} & 1878.4 & \textbf{20.9}\\
     6 & 20 & 494.4 & \textbf{11.3} & 1028.4 & \textbf{16.3} & 2153.0 & \textbf{37.1}\\
     & 10 & 643.6 & \textbf{9.6} & 1362.6 & \textbf{16.2} & 2785.5 & \textbf{50.4}\\
     & 5 & 874.6 & \textbf{14.4} & 1724.7 & \textbf{27.9} & 3603.8 & \textbf{69.7}\\
    \midrule
     & 100 & -- & -- & 518.6 & \textbf{4.5} & 1897.7 & \textbf{18.4}\\
     & 50 & 275.8 & \textbf{2.8} & 907.5 & \textbf{10.1} & 2131.2 & \textbf{19.4}\\
     7 & 20 & 557.8 & \textbf{6.4} & 1215.3 & \textbf{12.0} & 2571.2 & \textbf{24.3}\\
     & 10 & 691.2 & \textbf{8.8} & 1543.1 & \textbf{18.0} & 3186.0 & \textbf{36.1}\\
     & 5 & 909.6 & \textbf{16.2} & 1915.4 & \textbf{33.5} & 4086.0 & \textbf{66.5}\\
    \bottomrule
    \end{tabular}
    \label{tab:convergence_runtime_appendix}
\end{table}

\subsection{\rebuttal{Computational efficiency of the B-spline Optimization}}\label{sec:computational_efficiency_appendix}
\rebuttal{To demonstrate the computational efficiency of our B-spline implementation, we conducted an ablation study on the convergence time of the B-spline optimization problem. 
The results, presented in Table~\ref{tab:computational_efficiency_bspline_algorithm}, are based on a sequence from the NeuroBEM dataset, where the drone follows a random trajectory, split into sequential, non-overlapping segments of 0.1, 0.2, 0.5, and 1.0 s. 
We evaluated different B-spline orders, control point spacings, and torque measurement rates, all within the practical operating range of \method. 
Control points were initialized via linear interpolation of gyroscope measurements. 
The experiments are run on a laptop running Ubuntu 20.04 and equipped with an Intel Core i9 2.3 GHz CPU.
The results confirm that our B-spline optimization is highly efficient.}
\begin{table}[!ht]
    \centering
    
    \caption{\rebuttal{Convergence time in [ms] of the optimization problem that optimizes the B-spline control points according to the quadrotor's rotational dynamics model (see Eq.~\ref{eq:bspline_meas_model}). We selected a sequence from the NeuroBEM dataset where the drone follows a random trajectory. This trajectory is split into sequential, non-overlapping segments. We evaluated different B-spline orders, control point spacings, and torque measurement rates, all within the practical operating range of \method. We report the average convergence time computed over all the trajectory segments. -- indicates an infeasible configuration caused by a trajectory segment that is too short relative to the chosen B-spline order and control point spacing.}
} 
    \setlength{\tabcolsep}{4.5pt}
    \renewcommand{\arraystretch}{1.2}
    \begin{tabular}{c|c|cccc|cccc}
    \toprule
    \textbf{B-spline} & \textbf{Ctrl. Points} & \multicolumn{8}{c}{\textbf{Input Measurements Frequency} [Hz]} \\[2pt]
    \textbf{Order} & \textbf{Spacing} & \multicolumn{4}{c|}{100} & \multicolumn{4}{c}{200}\\[2pt]
    & [ms] & \multicolumn{8}{c}{\textbf{Trajectory Segment Length} [s]}\\
    &  & 0.1 & 0.2 & 0.5 & 1.0 & 0.1 & 0.2 & 0.5 & 1.0\\
    \midrule
     & 50 & -- & -- & 0.6 & 1.7 & -- & -- & 0.9 & 2.5\\
     5 & 20 & -- & 0.5 & 1.5 & 3.2 & -- & 0.3 & 1.0 & 2.0\\
     & 10 & 1.0 & 1.9 & 5.0 & 10.1 & 0.4 & 1.0 & 2.1 & 4.3\\
    \midrule
     & 50 & -- & -- & 0.8 & 2.7 & -- & -- & 1.2 & 3.2\\
     6 & 20 & -- & 0.6 & 2.3 & 5.0 & -- & 0.7 & 2.7 & 5.3\\
     & 10 & 0.8 & 1.9 & 6.8 & 11.0 & 0.6 & 1.7 & 4.3 & 9.3\\
    \midrule
     & 50 & -- & -- & 1.0 & 3.1 & -- & -- & 1.3 & 4.2\\
     7 & 20 & -- & 0.7 & 3.3 & 7.0 & -- & 0.8 & 3.2 & 6.7\\
     & 10 & 0.6 & 1.6 & 4.7 & 9.5 & 0.8 & 2.1 & 5.7 & 12.0\\
    \bottomrule
    \end{tabular}
    \label{tab:computational_efficiency_bspline_algorithm}
\end{table}

%% Use plainnat to work nicely with natbib. 
{\bibliographystyle{unsrtnat}
\bibliography{references}}
\clearpage

\begin{IEEEbiography}[{\includegraphics[width=1in,height=1.25in,clip,keepaspectratio]{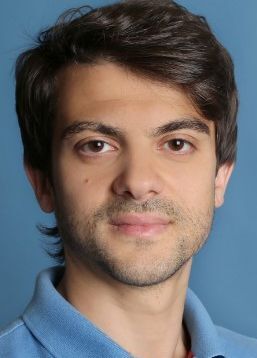}}]{Giovanni Cioffi} %
holds an M.Sc. in Mechanical Engineering from ETH Zürich, Switzerland, which he obtained in 2019. He is currently pursuing a Ph.D. at the University of Zürich under the supervision of Prof. Davide Scaramuzza. His research centers on the intersection of computer vision and robotics, exploring topics such as visual(-inertial) odometry and SLAM. His contributions were recognized by multiple awards in top-tier robotic conferences and journals, such as the IROS 2023 Best Paper Award and the RA-L 2021 Best Paper Award.
\end{IEEEbiography}
\begin{IEEEbiography}[{\includegraphics[width=1in,height=1.25in,clip,keepaspectratio]{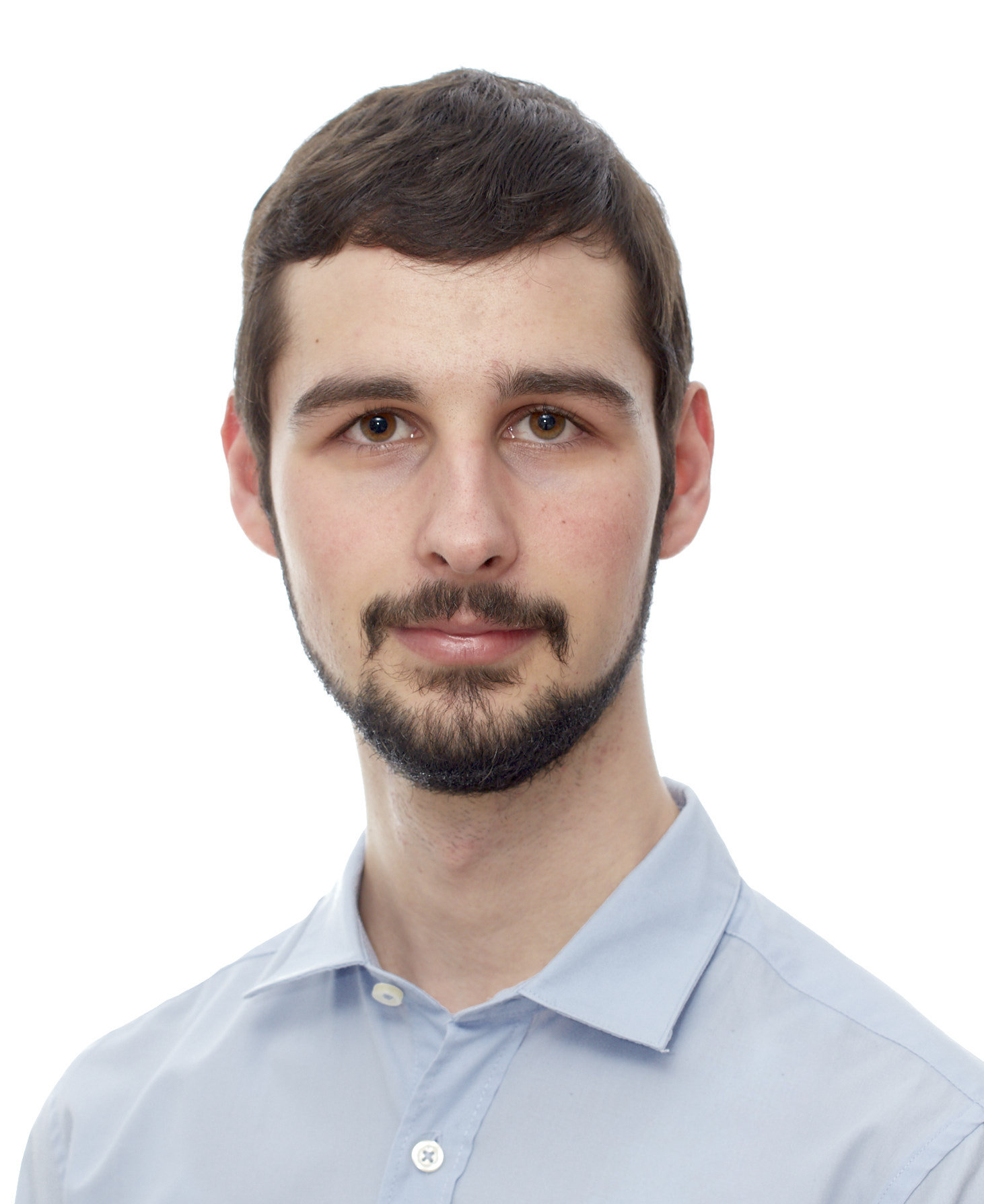}}]{Leonard Bauersfeld} received his M.Sc. degree in robotics, system and control from ETH Zurich, Switzerland
in 2020. He is currently a PhD student in the Robotics and Perception Group at the University of Zurich, led by Prof. Davide Scaramuzza. His research interests are autonomous vision-based quadrotor flight and quadrotor simulations. He works on novel approaches, combining first-principles methods with modern data-driven models to advance agile quadrotor flight. 
\end{IEEEbiography}
\begin{IEEEbiography}[{\includegraphics[width=1in,height=1.25in,clip,keepaspectratio]{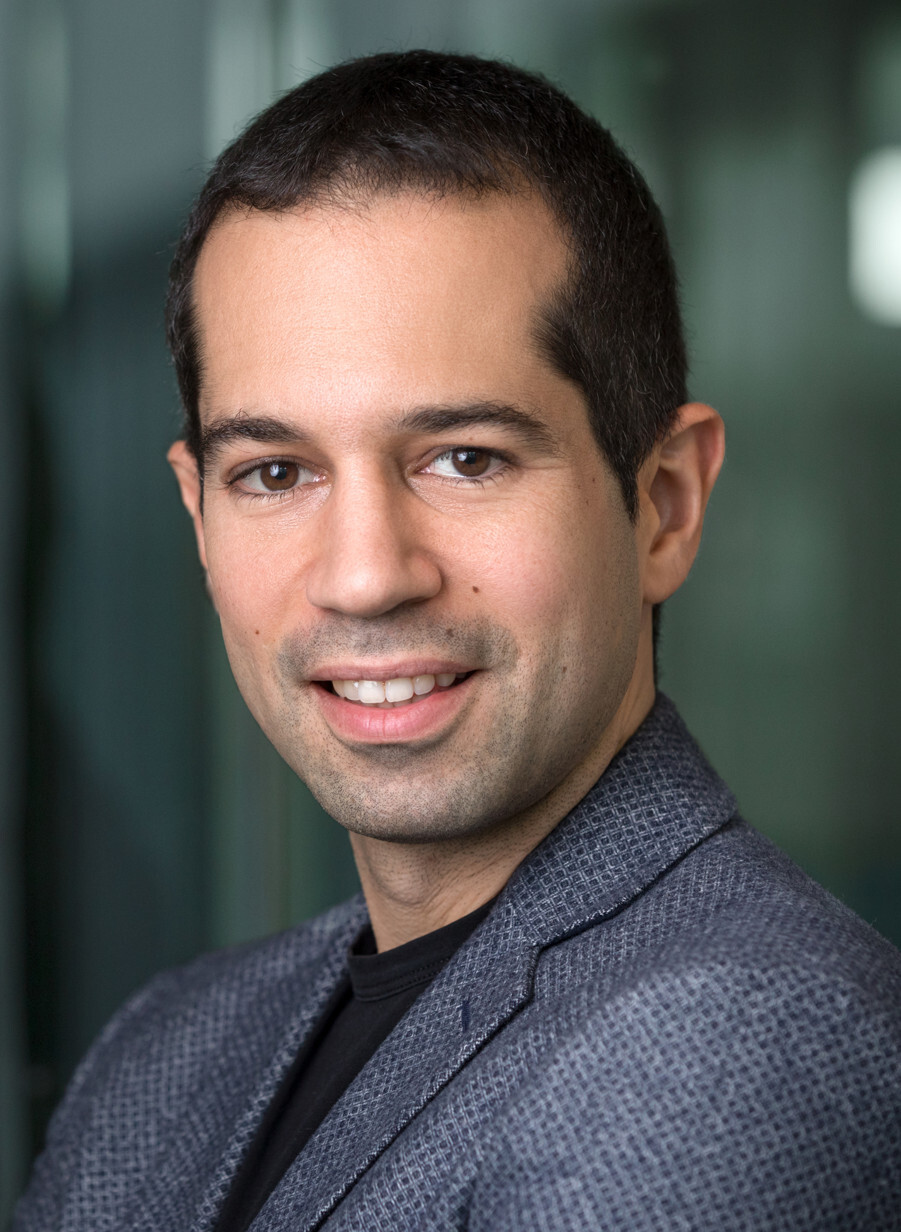}}]{Davide Scaramuzza} %
is a Professor of Robotics and Perception at the University of Zurich. He did his Ph.D. at ETH Zurich, a postdoc at the University of Pennsylvania, and was a visiting professor at Stanford University. His research focuses on autonomous, agile microdrone navigation using standard and event-based cameras. He pioneered autonomous, vision-based navigation of drones, which inspired the navigation algorithm of the NASA Mars helicopter and many drone companies. He contributed significantly to visual-inertial state estimation, vision-based agile navigation of microdrones, and low-latency, robust perception with event cameras, which were transferred to many products, from drones to automobiles, cameras, AR/VR headsets, and mobile devices. In 2022, his team demonstrated that an AI-controlled, vision-based drone could outperform the world champions of drone racing, a result that was published in Nature. He is a consultant for the United Nations on disaster response, AI for good, and disarmament. He has won many awards, including an IEEE Technical Field Award, the IEEE Robotics and Automation Society Early Career Award, a European Research Council Consolidator Grant, a Google Research Award, two NASA TechBrief Awards, and many paper awards. In 2015, he co-founded Zurich-Eye, today Meta Zurich, which developed the world-leading virtual-reality headset Meta Quest. In 2020, he co-founded SUIND, which builds autonomous drones for precision agriculture. Many aspects of his research have been featured in the media, such as The New York Times, The Economist, and Forbes.
\end{IEEEbiography}

\end{document}